\documentclass{article}

\usepackage[margin=1.1in]{geometry} % 
\usepackage[font={small, sf}]{caption}
\usepackage[hang,flushmargin]{footmisc} 
\usepackage{array}

\usepackage[utf8]{inputenc}

\usepackage{float}
\usepackage{empheq}
\usepackage{comment}
\usepackage[T1]{fontenc}
\usepackage{makecell}
\usepackage{booktabs}
\usepackage{amsfonts}
\usepackage{authblk}
\usepackage{hyperref}
\usepackage{cite}
\usepackage{enumitem}
\usepackage{wrapfig}
\usepackage{titlesec}
\usepackage{geometry}

\usepackage{subfiles} % Best loaded last in the preamble

\makeatletter
\newcommand{\thickhline}{%
    \noalign {\ifnum 0=`}\fi \hrule height 1pt
    \futurelet \reserved@a \@xhline
}
\newcolumntype{"}{@{\vrule width 1pt}}
\makeatother
\newcommand*\pFq[2]{{}_{#1}F_{#2}\genfrac[]{0pt}{}}

\titlespacing{\section}{0pt}{\parskip}{-\parskip}

\titlespacing{\subsection}{3pt}{\parskip}{-\parskip}
\title{Algorithm-assisted discovery of an intrinsic order among mathematical constants}

\author{Rotem Elimelech}
\author{Ofir David}
\author{Carlos De la Cruz Mengual}
\author{Rotem Kalisch}
\author{Wolfgang Berndt} 
\author{Michael Shalyt}
\author{Mark Silberstein}
\author{Yaron Hadad}
\author{Ido Kaminer}

\affil{Technion - Israel Institute of Technology, Haifa 3200003, Israel}
\affil[]{Corresponding author: kaminer@technion.ac.il}

\date{}

\begin{document}

\maketitle

\section*{Abstract}
In recent decades, a growing number of discoveries in fields of mathematics have been assisted by computer algorithms, primarily for exploring large parameter spaces that humans would take too long to investigate. 
% First citations - algorithm-driven math 
%\cite{graffiti,every_planar_map_is_four_colorable,zeilberger1990holonomic,zeilberger1990fast,wilf1992algorithmic,TRM,fawzi2022_deepmind_matmul}
As computers and algorithms become more powerful, an intriguing possibility arises---the interplay between human intuition and computer algorithms can lead to discoveries of novel mathematical concepts that would otherwise remain elusive. 
% Second citation - broader vision of AI4math
%\cite{wang1960toward, davis1982knowledge,lenat1984_why_AM_EURISKO_work,wolfram2002new,buchberger2006theorema,davies2021advancing}. 
To realize this perspective, we
%In this study, we showcase the validity of this concept. We
have developed a massively parallel computer algorithm that discovers an unprecedented number of continued fraction formulas for fundamental mathematical constants. 
The sheer number of formulas discovered by the algorithm unveils a novel mathematical structure that we call the \textit{conservative matrix field}. Such matrix fields (1) unify thousands of existing formulas, (2) generate infinitely many new formulas, and most importantly, (3) lead to unexpected relations between different mathematical constants, including multiple integer values of the Riemann zeta function.
Conservative matrix fields also enable new mathematical proofs of irrationality. In particular, we can use them to generalize the celebrated proof by Apéry for the irrationality of $\zeta(3)$. %\cite{apery,poorten}.
Utilizing thousands of personal computers worldwide%\cite{boinc}
, our computer-supported research strategy demonstrates the power of experimental mathematics, highlighting the prospects of large-scale computational approaches to tackle longstanding open problems and discover unexpected connections across diverse fields of science.

\section{Mathematical constants}\label{sec:intro}
Mathematical constants emerge naturally in various fields of science and branches of mathematics, such as geometry, combinatorics, number theory, and  probability theory \cite{finch2003mathematical}. 
Well-known examples of constants are the ratio between the circumference and the diameter of a circle $\pi$, the base of the natural logarithm $e$, the golden ratio $\varphi$, and the values of the infinite sum for integers $n \geq 2$:
\begin{equation*} \label{eq:def_zeta}
	\zeta(n) \coloneqq \frac1{1^n} + \frac1{2^n} + \frac1{3^n} + \cdots
\end{equation*}
The latter are the integer values of the Riemann zeta function $\zeta$, a cornerstone of mathematical exploration in the field of number theory, due to the remarkable connection between its complex zeros and the asymptotic distribution of prime numbers. 
The precise distribution of these zeros of $\zeta$ constitutes the Riemann hypothesis, arguably the most important unsolved question in pure mathematics to this day.

The occurrence of a constant in diverse mathematical contexts reflects its significance and often leads to the discovery of fundamental connections. A remarkable example is given by the so-called Basel problem, posed in 1650, yet only solved in the mid-18th century by Euler \cite{euler_original_zeta} (see also \cite{euler_and_the_zeta_function}). His solution established the relation $\zeta(2) = \pi^2/6$
and, in general, expressed all positive even values $\zeta(2n)$ as rational multiples of even powers of $\pi$. Euler's discovery also hinted for the first time at the profound, yet \emph{a priori} not obvious, relevance of $\pi$ in questions related to the distribution of prime numbers. 
Notably, no analogous relation has been found for the odd values $\zeta(2n+1)$ to date, three centuries after Euler's work. 

Despite their tremendous impact, discoveries of new relations between mathematical constants and other mathematical structures are sporadic events that stem from strong intuitive leaps or great strokes of creativity. In light of this, a captivating possibility is that a deeper underlying concept exists, encompassing all or a significant group of constants and providing a framework for classifying and ordering them. 
Here we propose such a mathematical structure---called \textit{conservative matrix field}---and show that it not only recreates known relations between constants, but also reveals new ones. 
Through these relations, the matrix fields help to order constants within a cohesive framework that may shed new light on their interrelationships, intrinsic properties, and complexity.

\subsection{The complexity of mathematical constants}
Central to understanding the intrinsic nature of a mathematical constant is the question of its irrationality, which can be understood as a measure of its complexity. A rational number can be regarded as possessing the most simple representation, being expressible as a quotient of two integers. In turn, an irrational number exhibits a higher level of complexity, for it cannot be represented by such a quotient.
%but can only be described as the limit of a sequence thereof. 

Even though the question of whether a number is rational or irrational may appear innocent at first glance, for numerous fundamental constants, determining their rationality can be a highly non-trivial task. The situation for the odd values of the Riemann zeta function serves to illustrate this claim. Indeed, the proof of the irrationality of $\zeta(3)$ by Ap\'ery in 1978 \cite{apery,poorten} stands as the only definitive accomplishment in this regard, while the matter remains unsettled for all of the remaining odd values $\zeta(2n+1)$; see \cite{ball2001irrationalite,zudilin_one_zetas_is_ir,Zudilin_z5_ir_measure} for partial results in this direction. The odd values of the zeta function, along with the Catalan, Gompertz, and Euler--Mascheroni constants, represent a large collection of prominent mathematical constants for which the question of irrationality is still open.

The distinction rational vs. irrational offers a first kind of hierarchy that orders numbers according to their complexity. An attempt to refine this binary hierarchy arises from the concept of \textit{irrationality measure} \cite{intro_num_theo_ir_measure}, a numerical value that quantifies how fast a constant may be approximated by an infinite sequence of distinct rational numbers. In this sense, any number that admits a fast enough rational approximation must be irrational. Rational numbers are characterized by an irrationality measure of zero, while irrational numbers are characterized by an irrationality measure of at least one. Although there are numbers with arbitrary irrationality measures greater than one, it is noteworthy that \emph{most} real numbers possess an irrationality measure of one \cite{Khinchin}. Consequently, there is more to be desired from a further-refined dichotomy between rational and irrational numbers.

Our endeavor to establish a finer hierarchy revolves around quantifying the complexity of the actual formulas that may represent each number, and clustering them accordingly. We now explain our notion of formulas and elaborate on why this specific class of formulas is highly suitable for computer-supported research.

\subsection{Algorithm-driven discovery of formulas for mathematical constants}
Ap\'ery's celebrated proof of the irrationality of $\zeta(3)$ relies on producing an efficient approximation of this number via the continued fraction formula
\begin{equation} \label{eq:apery_pcf}
\frac{6}{\zeta(3)} = 5 - \cfrac{1^6}
               {117 - \cfrac{2^6}
                   {535 - \cfrac{3^6}
                        {\begin{array}{ll}
                       \ddots \\[-10pt]
                       & \hspace{-4pt}-\, \cfrac{\textstyle n^6}{\textstyle 17(n^3 + (n+1)^3) - 12 (2n+1)- \ddots}
                        \end{array}}}}
\end{equation}
Truncating such an infinite continued fraction at a finite number of steps yields a sequence of rational numbers approximating the target constant. In this way, the exploration of continued fractions offers an approach for estimating irrationality measures. 

Within the class of continued fractions, formulas like Eq. \ref{eq:apery_pcf} constitute the distinguished subclass of \emph{polynomial continued fractions}, for which the partial numerators and denominators are defined by polynomials with integer coefficients as a function of the \emph{depth} $n$. The study of polynomial continued fractions is particularly attractive due to their broad mathematical scope and computational simplicity \cite[\S VI]{Perron2} (see also \cite{bowman2002polynomial,laughlin2005real,handbook_of_cfs,zeta5_dzmitry}). Remarkably, these continued fractions encompass a broad range of mathematical functions including trigonometric, Bessel, gamma, and hypergeometric functions; they represent a large collection of constants \cite{endnote_formula_is_pcf}; 
they generalize a wide class of infinite sums \cite{euler1748}; and they represent linear recurrence formulas with polynomial coefficients \cite{Perron2}. The computational simplicity of polynomial continued fractions stems from two key aspects. Firstly, the space of polynomial continued fractions is countable, which enables its systematic exploration and analysis. Secondly, the rational approximation by a polynomial continued fraction can be evaluated very efficiently \cite[\S 5.2]{Numerical_recipes}. Those two facts make this space a fitting candidate for computer-supported research. 
% maybe add : \cite{alan_turing_computable_numbers}

The Ramanujan Machine project \cite{TRM}, named to honor Srinivasa Ramanujan's unique contributions to mathematics \cite{ramanujan_notebooks}, proposed to automate the process of discovery of formulas via an algorithmic approach. The project developed algorithms to find formulas for constants numerically, yielding many new ones such as notable formulas for Catalan's constant. The most successful algorithm employed there relied on a brute-force search over the space of polynomial continued fractions.

To broaden the exploration beyond the space that can be covered by brute-force searches, 
one is confronted with the necessity to develop a more sophisticated exploration strategy. 
It was through the analysis of the database of formulas obtained in \cite{TRM} that we were able to identify an intriguing numerical property in the space of formulas. This property, called \emph{factorial reduction} (defined in Section \ref{ssec:lab_and_fr}), guided the development of a more efficient algorithm for formula discovery. 
%The property of factorial reduction is extremely rare in the space of all arbitrary continued fraction formulas, yet it is satisfied by all the formulas obtained for the constants discussed here. 
%In fact, even more surprisingly, factorial reduction is also satisfied by the continued fraction formulas and infinite sums found over the centuries for these constants (e.g., \cite{Perron2,laughlin2005real,handbook_of_cfs,bowman2002polynomial}).

Our novel algorithm is thus designed to search for formulas based on factorial reduction, 
discovering hundreds of polynomial continued fraction formulas for well-known mathematical constants.
This output significantly exceeds what has been manually discovered in the past (e.g., \cite{handbook_of_cfs}).
%, well beyond what was discovered by humans (e.g., \cite{handbook_of_cfs}).
%produced a much greater number of formulas than any previous work. 
The algorithm is heuristic in nature, as it prunes the search space based on the \textit{conjecture} that continued fractions representing known constants tend to have factorial reduction. At any rate, 
the actual formulas that the algorithm discovers can be and are independently verified.
A search utilizing the factorial reduction property makes the algorithm conducive to distributed computing, enabling us to implement a massively parallel search, with which we covered a significantly larger space of formulas than what was possible before.

\begin{figure}[H]%[t!]
    \vspace{-0.2cm} 
    \centering
    \includegraphics[width=12cm]{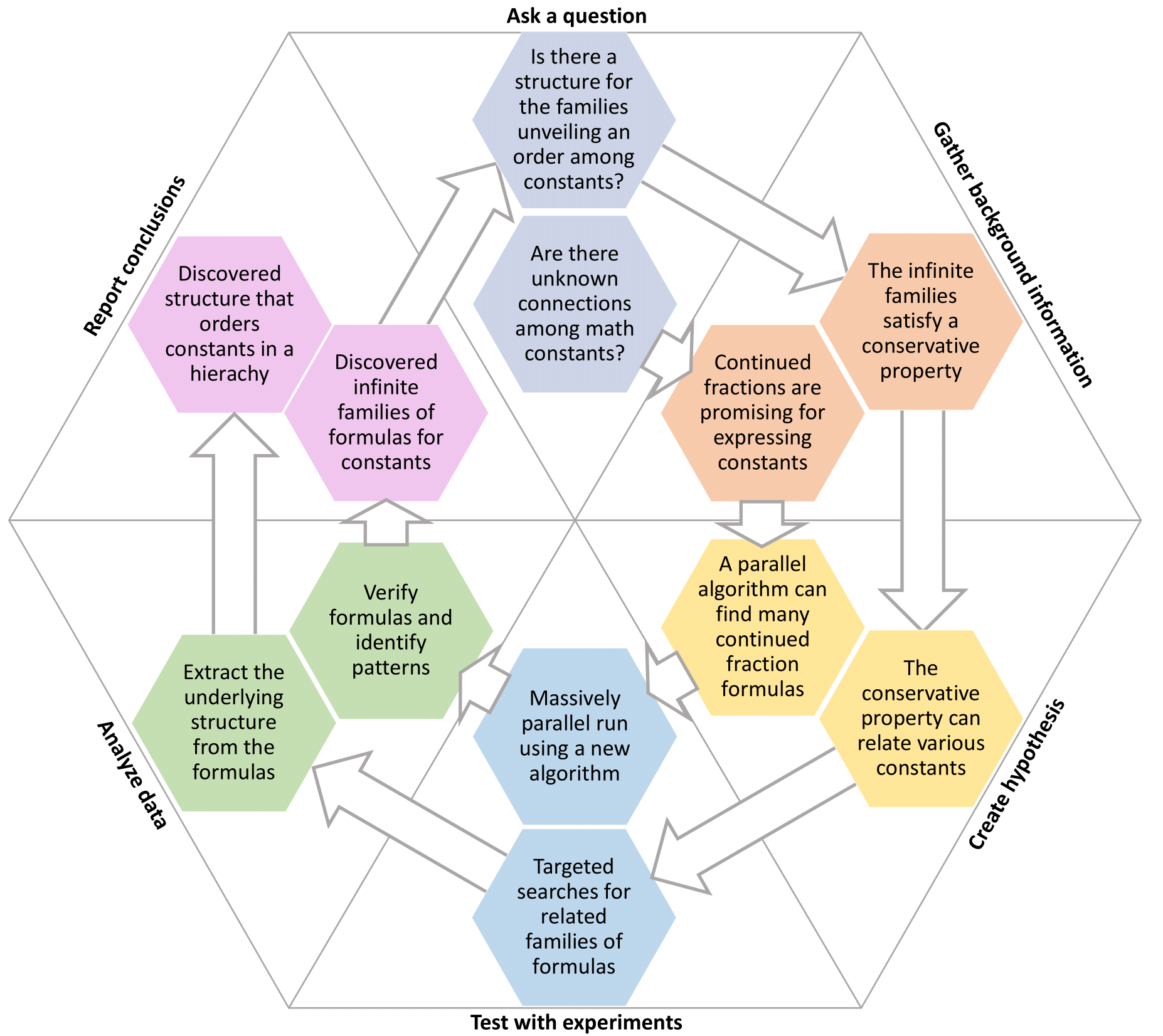}
    \vspace{-0.4cm}
    \caption{\textbf{Algorithmic-driven scientific method in experimental mathematics and its realization in our work}. Our work encompasses two cycles of the scientific method, leading to the discovery of a novel mathematical concept.}
    \vspace{-0.2cm}
    \label{fig:scientic_method}
\end{figure}

Our algorithm-driven approach exemplifies the vision of \textit{experimental mathematics} \cite{A_eq_B_Zeilberger,wolfram2002new,buchberger2006theorema,bailey2007experimental}---leveraging algorithms at scale to generate new insights in mathematics (see, e.g., \cite{graffiti,
%every_planar_map_is_four_colorable,
%zeilberger1990holonomic,
%zeilberger1990fast,
%wilf1992algorithmic,
TRM,
%wang1960toward, 
%davis1982knowledge,
%lenat1984_why_AM_EURISKO_work,
%wolfram2002new,
%buchberger2006theorema,
davies2021advancing,
fawzi2022_deepmind_matmul}).
Through the generation of large amounts of data, algorithms offer valuable hints about patterns and behaviors of mathematical structures that would be otherwise imperceptible within a reasonable timeframe. These hints guide mathematicians in formulating stronger conjectures and even provide a blueprint for rigorous proofs and generalizations. This process can be regarded as a cycle: the results of algorithmic experiments create new conjectures that themselves inspire new algorithmic experiments, leading to further discoveries and a deeper understanding of the original problem. 
Our work executes two cycles of an algorithmic-driven scientific method (Fig. \ref{fig:scientic_method}),
leading to the discovery of an underlying model---a novel mathematical structure.

We see a parallel between the relationship of algorithms to mathematics and that of experiments to theoretical physics. Computational frameworks can serve as virtual laboratories for mathematicians, enabling systematic exploration of mathematical hypotheses on a large scale. The greater computational power available, the more precise and comprehensive the experiments, providing potential breakthrough opportunities. This collaborative approach between algorithms and experts facilitates the exploration and testing of hypotheses, leading to an accelerated understanding of unsolved problems and augments the serendipitous insights of brilliant mathematicians.

\subsection{Discovery of a novel mathematical structure: conservative matrix field}

The insight that led us to find the concept of conservative matrix fields resulted exclusively from having the large dataset of polynomial continued fraction formulas discovered by our novel algorithm \cite{TRM_RESULTS}.
When examining these formulas, certain recurring symmetries and patterns arise. 
As an example, note the resemblance between the following three formulas for $\pi$:
    \vspace{-0.2cm}
\begin{equation} \label{eq:similar_pi_pcfs}
    \centering 
    \includegraphics[width=13.5cm]{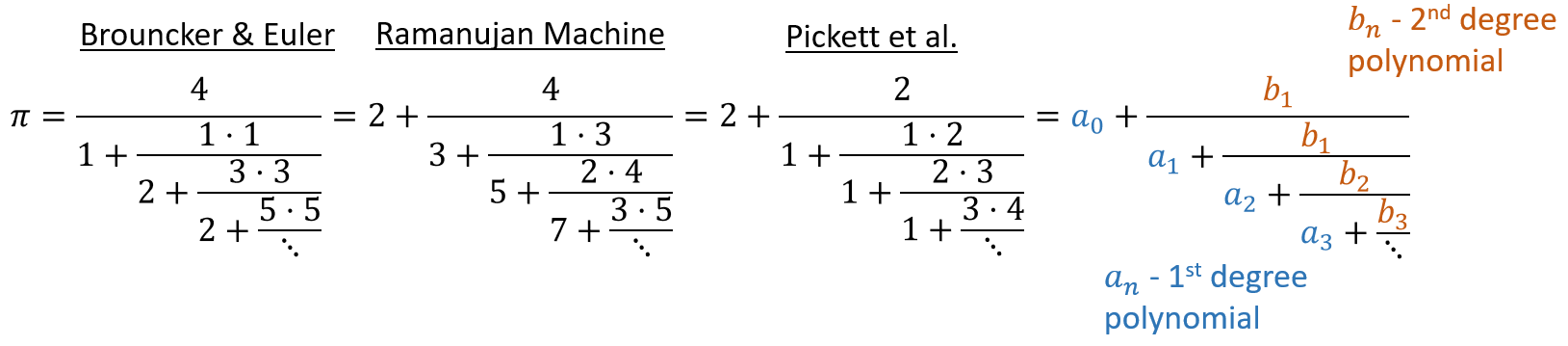}
    \vspace{-0.4cm}
\end{equation}
The first one was discovered by Brouncker and proven by Euler
(see \cite{hist_of_pi}), the second one by the Ramanujan Machine \cite{TRM}, and the third one by Pickett et al. \cite{pickett2008another}.
Such striking similarities were found in many more examples in our dataset of formulas, sparking the question of whether there exists a governing structure within the space of formulas for each given constant. 
%To gather the data that helped finding this governing structure, we developed and implemented
%a powerful distributed algorithm.

The massive number of formulas produced by our algorithm allowed us to identify several infinite parametric families of formulas representing constants such as $e,\,\ln(2),\,\pi$ and $\zeta(3)$ (see Section \ref{sec:infinite_families}). 
A study of the relationships between the formulas revealed a mathematical structure, coined conservative matrix field (Section \ref{sec:cmf}), which unifies the infinite families and even derives new formulas with faster rates of convergence. This structure was identified by observing a kind of ``conservation law'', reminiscent of the defining property of conservative vector fields in physics. 

A remarkable application of conservative matrix fields is identifying promising paths for proving the irrationality of constants, as we exemplify for $\zeta(3)$ \cite{CMF}.
This approach now directly applies to more cases (Appendix I%\ref{app:ir_meas_cmf_family}
).

We put forward the conjecture that these matrix fields have the potential to unify all polynomial continued fraction formulas associated with a specific constant. 
Interestingly, matrix fields also unveil connections between different constants, for instance, between $\pi^2$ and Catalan's constant, $\pi^3$ and $\zeta(3)$, or $e$ and Gompertz's constant. Based on these findings, we hypothesize that many constants, including values of the Riemann zeta function, can be interconnected through the framework of conservative matrix fields, supporting the idea that they give rise to a new hierarchy of mathematical constants.

\section{The Distributed Factorial Reduction algorithm} \label{sec:the_algorithm}

Our algorithm aims to discover formulas that equate linear fractional transformations of mathematical constants and polynomial continued fractions, both with integer coefficients.
Due to the nature of this problem, the search space is intrinsically infinite. Even if we limit our search to a small list of constants, the search space encompasses all their combinations with arbitrary coefficients, and the number of combinations grows exponentially with the maximal coefficients allowed.
A greater difficulty lies in the fact that an attempt to explore the search space in parallel by distributing it across different workers leads to redundant computations of the same expressions. Specifically, 
division of the search space between workers over the space of constants requires repeated computing of the same continued fraction by different workers, each equating the result to different constants or different transformations of constants.
Moreover, sharing the already computed results among workers to avoid such redundancy leads to a prohibitively high communication overhead between them.

Our algorithm overcomes these challenges by identifying continued fractions that possess a novel property---which we call \textit{factorial reduction}. This identification enables us to pinpoint which continued fractions relate to (well-known) mathematical constants without having to equate them to the (infinite) space of possible constants and transformations of constants.
Our algorithm thus avoids the need to scan the space of constants, which substantially reduces its complexity, and most importantly, makes it amenable to large-scale distributed computing.
This way, we scan the search space across non-communicating workers without causing redundant calculations and avoid the communication bottleneck. 

We conjecture that the property of factorial reduction is a signature of recurrence sequences that converge to mathematical constants. Consequently, our algorithm not only searches for conjectured formulas but is itself based on a conjecture. Importantly, every formula generated by the algorithm is independently verified, thereby liberating it from the necessity of justifying the factorial reduction algorithm, making it a stand-alone conjecture awaiting formal proof.

This conjecture-based algorithm proved to be extremely effective, leading to the discovery of hundreds of new formulas for mathematical constants, many of which would have been hard or impossible to discover by other means. Identifying this conjectured property was enabled by the abundance of formulas discovered by the older algorithms of the Ramanujan Machine project \cite{TRM_RESULTS}.
In retrospect, the substantial number of independently verified formulas exhibiting factorial reduction reinforces the belief that factorial reduction is a distinctive feature of formulas converging to constants of interest.

\subsection{Factorial reduction: an observation from the `lab' of experimental mathematics} \label{ssec:lab_and_fr}
At the core of our algorithm lies a novel observation about the greatest common divisors (GCDs) $g_n$ of the numerators $p_n$ and denominators $q_n$ of the convergents of polynomial continued fractions. The convergents are the rational numbers $p_n/q_n$ found by truncating an infinite continued fraction. More precisely, if $a_n$ and $b_n$ are the partial denominators and numerators of a continued fraction respectively (as in the rightmost term of Eq. \ref{eq:similar_pi_pcfs}), then \vspace{-4pt}
\begin{equation} \label{formula_structure}
    \frac{p_n}{q_n} = 
    a_0 + \cfrac{b_1}
        {a_1 + \cfrac{b_2}
            {\overset{\displaystyle a_2\,  + }{\phantom{a}} \ddots + \cfrac{b_n}
                {a_n}}}\vspace{-5pt}.
\end{equation}
These $p_n$ and $q_n$ are defined by the recursion $u_n = a_n u_{n-1} + b_n u_{n-2}$ with initial conditions $p_{-1}=1, p_0=a_0$ and $q_{-1}=0, q_0 = 1$.
The growth rate of the resulting convergent numerator $p_n$ and denominator $q_n$ is a power of a factorial, i.e. $p_n,q_n \sim (n!)^d$ for some positive integer $d$. See \cite{Wall_CF,Perron1,Perron2,TRM} for additional background on continued fractions.
 
Our algorithm relies on the observation that polynomial continued fractions that converge to mathematical constants exhibit a peculiar behavior: Denoting $g_n=GCD(p_n,q_n)$ as the greatest common divisor of the convergents, the \textit{reduced} numerator and denominator grow \textit{exponentially} at most, 
i.e. 
\begin{equation}\label{eq:s_definition}
    \frac{p_n}{g_n}, \frac{q_n}{g_n} \sim s^n,
\end{equation} 
instead of the much faster factorial growth rate of $p_n$ and $q_n$ before reduction. We name this observed property \textbf{factorial reduction}. See our complementary work \cite{FR} for additional details about this concept. 

Factorial reduction is found to be extremely rare: a wide search hints that the space of continued fractions with factorial reduction is a ``lower-dimensional'' subspace inside the space of all polynomial continued fractions. 
However, surprisingly, we observed factorial reduction in \textit{every} polynomial continued fraction formula found by the Ramanujan Machine using previous algorithms (see Appendix A) %\ref{app:trm_desc})
for a wide range of constants, including $\pi$, $\zeta(3)$, and Catalan's constant. 
%We have also identified this property in a large number of polynomial continued fractions and infinite sums converging to these constants that we found in literature spanning over hundreds of years (e.g., in 
In fact, we examined formulas converging to these constants in literature spanning over hundreds of years (e.g., \cite{
euler1748,euler_original_zeta,Perron1,
Perron2,bowman2002polynomial,laughlin2005real,handbook_of_cfs,ramanujan_notebooks
,naccache2022catalan,naccache2023balkans}).
Interestingly, all the continued fractions and infinite sums \cite{endnote_EulerCF} we checked displayed the property of factorial reduction.

\begin{figure}[H]%[b!]
    % \vspace{-0.5cm}
    \centering
    \includegraphics[width=15.5cm]{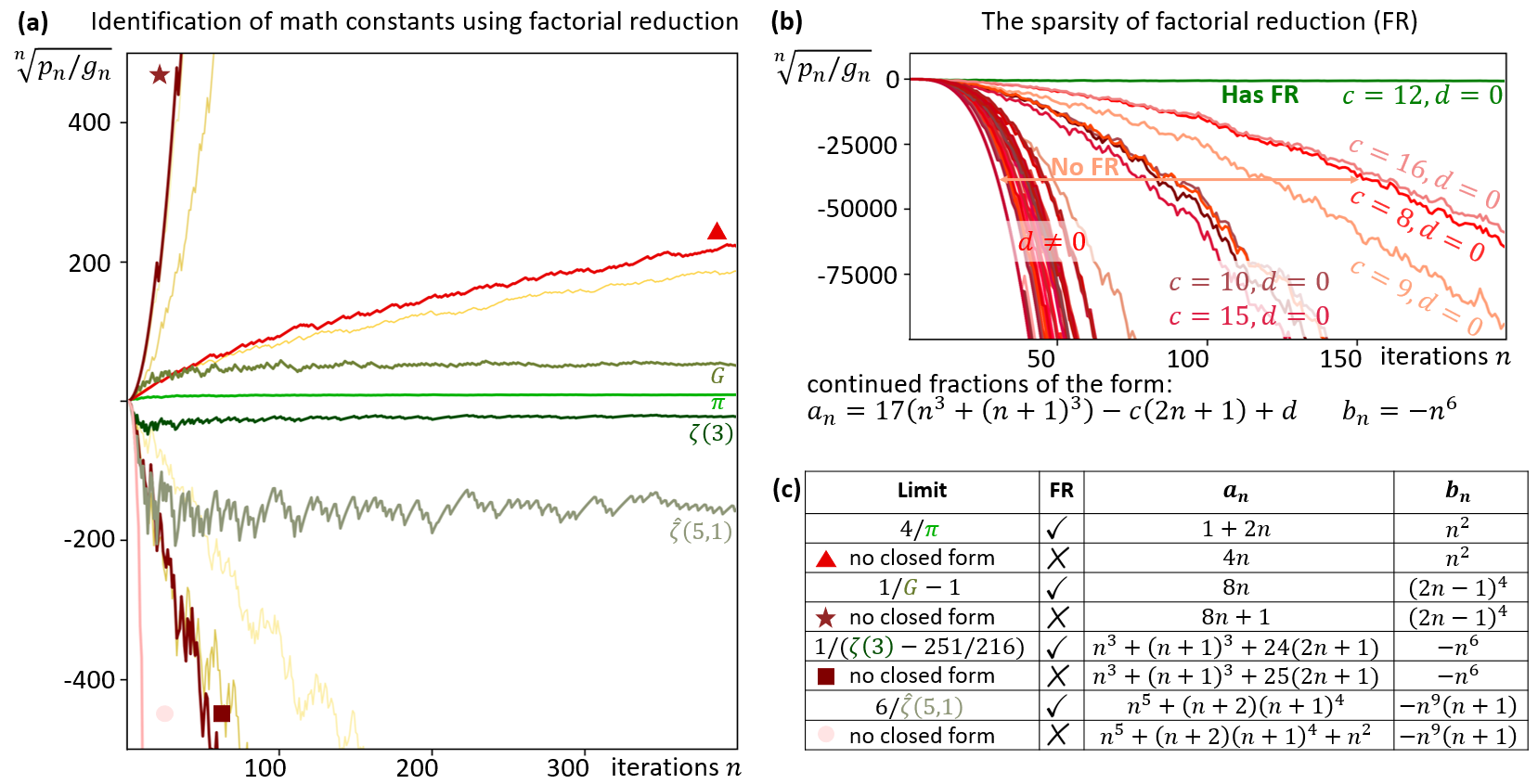}
    % \vspace{-0.5cm}
    \caption{\textbf{Observation of the factorial reduction (FR) property} - its connection to different constants and its sparsity in the wide space of continued fractions.
    (a) Calculation of the growth rate $s=\sqrt[n]{|p_n/g_n|}$ of the reduced convergents for different continued fractions, detailed in the table. We plot a negative $s$ value to denote a negative limit to $p_n/g_n$. Even very similar continued fractions can have vastly different growth rates. We observe exponential growth whenever the continued fraction converges to a mathematical constant. This approach identifies formulas for a wide range of mathematical constants. 
    (b) Comparison of the growth rate of continued fractions of the same form shows that the only one having factorial reduction is precisely the one found by Apéry \cite{apery,poorten} to converge to $\zeta(3)$. 
    The factorial reduction property is extremely rare, shown by its sensitivity to slight changes in the parameters of the continued fraction.
    (c) Table of the polynomial continued fractions calculated in panel (a), denoting which ones have or do not have factorial reduction, and which ones converge to a known constant or to a constant with no known closed form. Here, 
    $G$ is Catalan's constant, and
    $\hat\zeta(5,1)= \zeta(5)-\zeta(4)+\zeta(3)-\zeta(2)+1$ is the continued fraction from Eq. \ref{eq:zzz_general} with $s=5,R=1$. 
    }
    
    \label{fig:fr_s}
\end{figure}

The algorithm presented in this work relies on this unexpected, albeit simple property. Concretely, the algorithm numerically tests the growth rate of reduced convergents of polynomial continued fractions,
saving the ones that have exponential rates (rather than factorial rates).
This approach alleviates the need to identify a limit constant in advance. 
Thus, the property of factorial reduction enables efficient identification of continued fractions that converge to mathematical constants. 
Fig. \ref{fig:fr_s} displays the test for several polynomial continued fractions, showcasing the efficacy of our algorithm.

\vspace{1cm}

The concept of factorial reduction not only provides a useful identification tool for formulas of
constants but is also a mathematical curiosity on its own \cite{FR}. This concept is relevant to Ap\'ery-like irrationality proofs \cite{zeilberger_zudilin_irrationality}, and can help prove the irrationality of other constants (see also the notion of ``integer-ating factor'' in \cite{dougherty2022tweaking}). 

\subsection{Distributing the search for formulas} \label{ssec:dfr}
\begin{figure}[H]%[t!]
    \vspace{-0.5cm}  
    \centering
    \includegraphics[width=11.5cm]{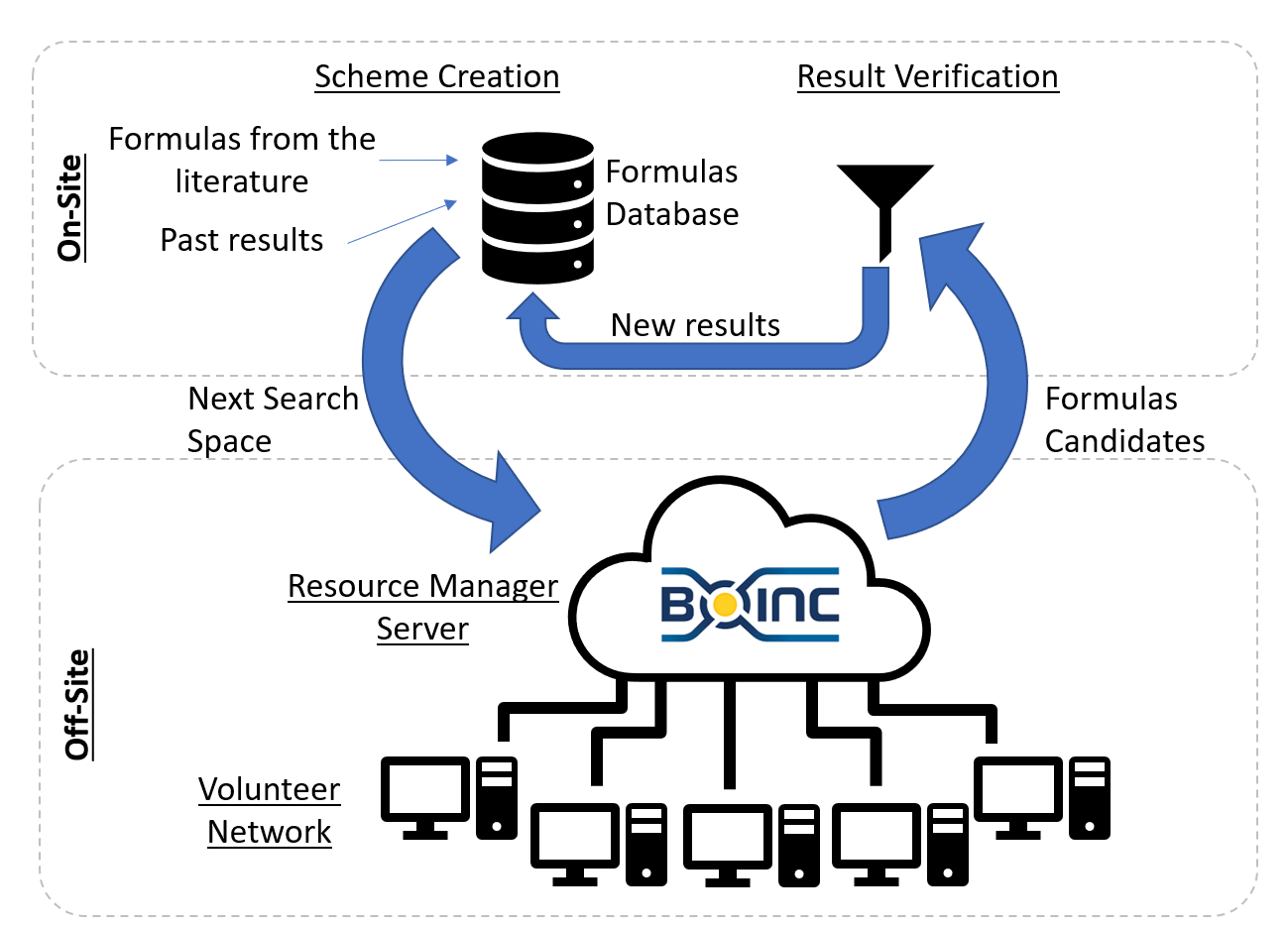}

    \caption{\textbf{Implementation of the Distributed Factorial Reduction (DFR) algorithm.}
    On-site scheme creation: Our database collects formulas from
    the literature and from past results of algorithms. This collection helps us define schemes of parameterized polynomial continued fractions. The range of parameters of the search space is decided based on our computational power. In this work, our search spaces are families of polynomial continued fractions.
    Off-site factorial reduction testing: Our BOINC resource manager server distributes the search space between the workers, who evaluate the polynomial continued fractions and test each one for factorial reduction. This step is  the most computationally intensive part of the algorithm. The computing power that enables this effort is donated by volunteers from the BOINC community, and, at the time of writing, involves over 5000 personal computers. 
    The rare events of positive identification are sent back for further analysis.
    On-site result verification: each identified continued fraction is first independently verified for factorial reduction. We then attempt to match each verified continued fraction to constants using PSLQ \cite{PSLQ1992,PSLQ1999}. Each PSLQ match is then validated to greater precision. We also collect the results for which a PSLQ match was not found, to enable future tests with additional constants, refined applications of PSLQ, and potentially superior algorithms that may emerge.}
    \label{fig:dfr}
\end{figure}

The identification of factorial reduction made possible the formulation of our Distributed Factorial Reduction (DFR) algorithm, whose key steps are (Fig. \ref{fig:dfr}):
\begin{itemize}
\item We create a scheme on-site that defines the search space of polynomial continued fractions.
\item This scheme is sent to a server, where it is parsed into smaller chunks, designed to compute in reasonable time on a standard home computer. The distribution of the computing chunks is done using the Berkeley Open Infrastructure for Network Computing (BOINC) \cite{boinc}. 
\item Each chunk is computed by off-site worker computers donated by volunteers, checking every polynomial continued fraction in the chunk for the factorial reduction property.
Each worker returns whether factorial reduction was identified (a very rare event). 
\item Every identified case of factorial reduction is verified automatically on-site.
%by calculating the continued fraction to a greater depth.
\item We then attempt to match each verified case to specific mathematical constants using the PSLQ algorithm \cite{PSLQ1992,PSLQ1999}. We save all the verified cases, with and without matching constants.
Each continued fraction for which matching constants were found is further validated by calculating it to a greater depth and comparing more digits, before storing the match as a new conjectured formula.
\end{itemize}

Although we have been able to identify constants for a significant majority of factorial reduction cases, we have encountered certain instances where a corresponding constant could not be identified. This is not surprising, as PSLQ is only able to test against the finite list of constants provided to it as an input.
Nevertheless, we also store as formula-candidates all the results for which PSLQ failed, i.e. found to have factorial reduction but without a connection to concrete constants (examples in Table A in Appendix C%\ref{table:sporadic_zetas} in Appendix \ref{app:all_pcfs}
).

The most intensive part of the search is the identification of factorial reduction, due to the large space of candidate polynomial continued fractions. 
For example, consider a search over polynomials $a_n,b_n$ of degrees 5 and 10 respectively.
This specific subspace of polynomial continued fractions is of interest because it was found to contain new formulas for $\zeta(5)$, a prominent constant for which the irrationality question remains open. 
For an intuitive estimate of the size of this search space, consider that even only testing polynomials with coefficients between -10 to 10 already contains $3\times10^{22}$ (74 bit) candidate continued fractions, an enormous effort for existing computational capabilities. We reduce the search space by selecting specific (non-canonical) representations of the polynomials, as described in Appendix C.2.
%\ref{app:ssec:zeta2_family}. 
Even this limited space requires immense computational resources.

Distributed computing is used in this part of the algorithm, to enable a search over such large spaces. We executed the distributed algorithm first on a Technion computing cluster (Zeus), and then distributed it on individual computers with the help of the BOINC community \cite{TRM_BOINC}. 
The thousands of new formulas discovered in this project are in large part thanks to the contribution of the BOINC community.
The choice of distributed computing via BOINC is especially valuable as it 
%creates the opportunity for the wider community 
allows everyone, novices or experts, the opportunity
to contribute to the discovery of new mathematics.

Through the support of the BOINC community, we are able to run our algorithm on extremely large search spaces, making it the largest computational experiment on formulas of mathematical constants.
Specifically, our Distributed Factorial Reduction algorithm 
has been running online since October 2021. At the time of writing, the algorithm is distributed over >5000 computers, and has been supported by >1000 volunteers throughout its operation.

\subsection{Example formulas discovered by our algorithm} \label{ssec:notable_results}
Our distributed algorithm led to the discovery of an unprecedented number of continued fraction formulas, creating a dataset that serves as the basis for the discoveries presented in this work.
This section presents selected examples of formulas from this dataset, 
all of which were unknown before this work, and most remain so far unproven.
A review of the results can be found on our website \cite{TRM_RESULTS}, and the full list is being prepared in an online library format \cite{lirec}, importable for future projects in experimental mathematics. 

We first present selected formulas for values of the Riemann zeta function. 
Our Distributed Factorial Reduction algorithm found results such as:
\begin{align}\label{eq:zeta2_family_member}
\frac{72}{72\zeta(2)-115} &= 21-\cfrac{1}{25-\cfrac{16}{
\ddots - \cfrac{n^4}{n^2 + (n + 1)^2+20 - \ddots}}}.
\end{align}
This formula and others can be generalized to infinite families that connect to the Lerch zeta function $\Phi$ (Appendix C).
%\ref{app:all_pcfs}). 
This function has explicit expressions for some of its values that combine more than one constant. For example, the following formula connects $\zeta(3)$ and $\pi^3$:
\begin{align}
\frac{9}{\Phi(1, 3,\frac{7}{3})} = 
\frac{9}{13\zeta(3) - \frac{1755}{64} + \frac{2\pi^3}{3\sqrt3}} &=
65 - \cfrac{81 \cdot 1^6}
  {249 - \cfrac{81 \cdot 2^6}
      {\ddots - \cfrac{81 \cdot n^6}
        { 9(n^3 + (n+1)^3) +56(2n+1) - \ddots}}}.
\label{eq:zeta3_family_intent_member}
\end{align}
The complexity of this formula and the examples below emphasize the prospects of the Distributed Factorial Reduction algorithm, since existing algorithms could not have discovered them.

Our algorithm discovered formulas that mix different zeta values, e.g., 
%\vspace{3pt}
\begin{equation} \label{eq:mixed_z3_z5}
\frac{1}{\zeta(3)-\zeta(2)+1} =\displaystyle 2 - \cfrac{2}{13 - \cfrac{96}{\ddots - \cfrac{n^5(n+1)}{ n^3 + (n+1)^3 + (n+1)^2 - \ddots}}}, \\
\end{equation}
\begin{equation*}
\frac{1}{\zeta(7)-4\zeta(3)+4} = 5 - \cfrac{1}{333 - \cfrac{16384}{\ddots - \cfrac{n^{14}}{n^7 + (n + 1)^7+8(n^5 + (n + 1)^5)-8(n^3 + (n + 1)^3)+4(2n + 1) - \ddots}}}.
\end{equation*}
The search also discovered infinite families of formulas that connect an arbitrary number of integer zeta values, with arguments up to half the degree of the polynomial $b_n$ in the continued fraction. We present one of these families, with polynomial degree $s$ and a root shifted by $R$, denoting its limit by $1/\hat\zeta(s,R)$: 
%\vspace{3pt}
\begin{equation} \label{eq:zzz_general}
\frac{1}{\zeta(s) + \alpha_{s-1} \zeta(s-1) + ... + \alpha_1 }=
1+R - \cfrac{1^{2s-1}(1+R)}
  {1+2^s+2^{s-1}R - \cfrac{2^{2s-1}(2+R)}
  	{\ddots - \cfrac{n^{2s-1}(n+R)}
      {n^s + (n+1)^s + R(n+1)^{s-1} - \ddots}}},
\end{equation}
where $\alpha_1, ..., \alpha_{s-1} \in \mathbb{Q}$ can be derived for every $R \in \mathbb{Q}$ (Appendix C.1).
%\ref{app:ssec:zzz}).
This general family of continued fractions is merely a subset of a larger infinite family that constructs a parametric set of $a_n$ polynomials for each $b_n$ with rational roots. 
We explore this family and provide a proof in Appendix C). %\ref{app:all_pcfs}.

Our algorithmic-driven research led to the discovery of formulas for other constants, such as the Catalan constant $G$
\begin{equation*}
\frac{1}{2G} = %\displaystyle 
1 - \cfrac{2}
  {7 - \cfrac{32}
%      {19 - \cfrac{162}
          {\ddots - \cfrac{2n^4}
            { 3n^2 + 3n + 1 - \ddots}}}.
\end{equation*}
The automated search led to many formulas for algebraic numbers, including ones with degrees greater than $2$, such as
\begin{equation*}
\frac{\sqrt[3]{4}+1}{\sqrt[3]{4}-1} =
5 - \cfrac{8}
  {15 - \cfrac{35}
%    {25 + \cfrac{-80}
      {\ddots - \cfrac{(3n)^2-1}
        {5 + 10n - \ddots}}}.
\end{equation*}
This continued fraction was also recently discovered in \cite{zeta5_dzmitry}, and is found here to be part of a parametric family of formulas for arbitrary-degree roots
\begin{equation*}
\label{eq:c1_pow_c}
\cfrac{\sqrt[c]{c+1}+1}{\sqrt[c]{c+1}-1}
=(2+c) - \cfrac{c^2-1}
  {3(2+c) - \cfrac{(2c)^2-1}
%    {5(2+c) - \cfrac{(3c)^2-1}
      {\ddots - \cfrac{(cn)^2-1}
        {(1+2n)(2+c) - \ddots}}}.
\end{equation*}
Appendix H %\ref{app:CMF_families} 
further shows that this parametric family is itself a particular case of a wider family of continued fractions whose limit is a quotient of hypergeometric functions ${}_2 \mathrm{F}_1$ \cite[\S 49]{Perron2}.

\section{Conservative matrix fields} \label{sec:cmf}
After implementing our Distributed Factorial Reduction algorithm, we seek to extract insights from the numerous resulting formulas. One such approach involves identifying clusters of formulas that exhibit common patterns and relate to distinct mathematical constants.

Remarkably, such clusters emerge naturally (see for example Fig. \ref{fig:lattice_motivation}), and exploring their properties leads to novel insights. Specifically, we present a mathematical structure that generalizes infinitely many polynomial continued fractions. This structure possesses intriguing mathematical properties and enables deriving additional polynomial continued fractions with their own unique characteristics. Notably, this structure leads to the polynomial continued fraction that Apéry employed to demonstrate the irrationality of $\zeta(3)$ and further enables the generalization of his approach to other constants.

\subsection{Infinite families of formulas found by the algorithm} \label{sec:infinite_families}
\begin{figure}[H]
    \centering
    \includegraphics[width=17cm]{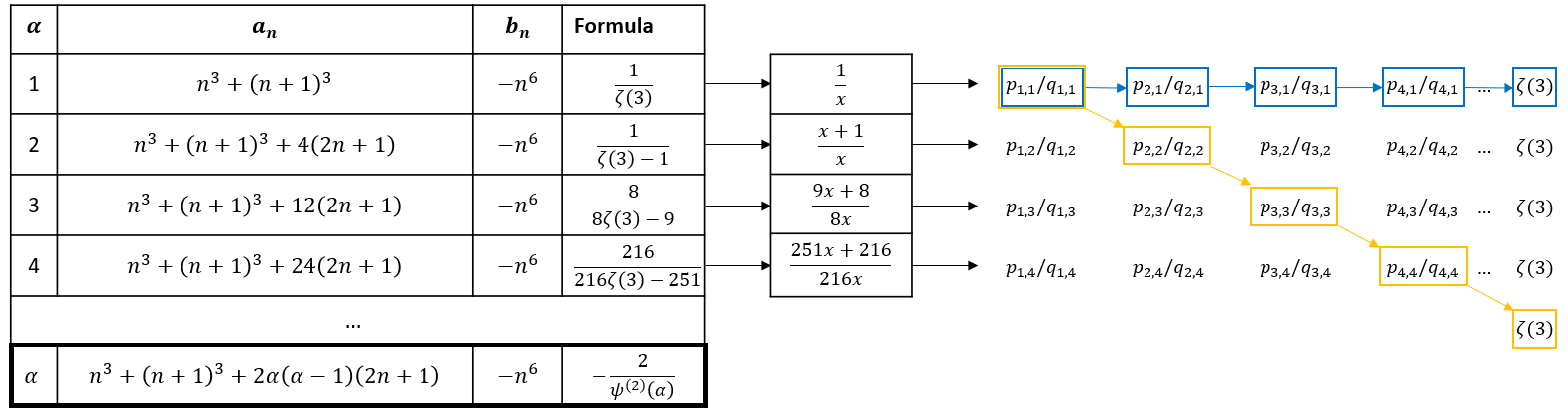}
    \caption{\textbf{Generating an efficient converging sequence from an infinite family of continued fractions}. The table presents a parametric family of continued fractions. The rows correspond to integer values of $\alpha$, which provide formulas converging to linear fractional transformations of $\zeta(3)$. Inverting this transformation for every element in each sequence creates sequences that all approach $\zeta(3)$. Then, we construct a new sequence by sampling the "diagonal" trajectory along the 2-dimensional grid of sequences, i.e., we select each consequent element from the consequent sequence. This constructed sequence creates an efficient approximation of $\zeta(3)$ that proves its irrationality.
    }
    \label{fig:lattice_motivation}
\end{figure}

The Distributed Factorial Reduction algorithm produces a multitude of formulas, which we group into infinite parametric families.
One illustrative example of an infinite family of formulas found by our algorithm is related to the constant $\zeta(3)$ (Fig. \ref{fig:lattice_motivation}). This family of formulas is parameterized by $\alpha$. For any rational value of $\alpha$, the limit was computationally observed to be a value of the polygamma function $\psi^{(2)}$.
Each integer value of $\alpha$ provides a formula for $\zeta(3)=-\psi^{(2)}(1)/2$ up to a linear fractional transformation. 

Applying the corresponding inverse transformation to each of the formulas for integer values of $\alpha$ yields an infinite family of sequences that all converge to $\zeta(3)$. We construct a new sequence of rational numbers that converge to $\zeta(3)$ by selecting the $n$-th element from the $n$-th sequence (illustrated in Fig. \ref{fig:lattice_motivation}). The new sequence exhibits a faster convergence rate than any of the continued fractions that constructed it, providing a more efficient representation of $\zeta(3)$. This sequence is exactly the one used by Apéry to prove the irrationality of $\zeta(3)$. 

The implication of this observation is that it is possible to create efficient formulas by building on infinite families of continued fractions. The next section formalizes this procedure, revealing a novel mathematical structure that generalizes our findings and even connects different mathematical constants.
Further examples of such infinite families are presented in Appendix C. %\ref{app:all_pcfs}.

\newpage

\subsection{Unifying formulas using the conservative matrix field}
\label{ssec:CMF}
We propose a mathematical structure that unites families of continued fractions representing a specific constant. To explain the origin and implications of this concept, we pass to a matrix representation of continued fractions and use the formulas of $\zeta(3)$ (Fig. \ref{fig:lattice_motivation}) as the leading example.

Being a composition of linear fractional transformations, any continued fraction can be represented as a multiplication of $2\times2$ matrices. The recurrence formula Eq. \ref{formula_structure} satisfied by the convergent numerators $p_n$ and denominators $q_n$ can be translated into:
\begin{equation} \label{eq:pcf_to_mat_mult}
\underbrace{\begin{pmatrix}
    p_{n-1} & p_{n} \\
    q_{n-1} & q_{n} 
\end{pmatrix}}_{V(n+1)} = 
\underbrace{\begin{pmatrix}
    1 & a_0 \\
    0 & 1 
\end{pmatrix}}_{V(1)} \cdot 
\underbrace{\begin{pmatrix}
    0 & b_{1} \\
    1 & a_{1} 
    \end{pmatrix}}_{M_X(1)} \cdot \ \cdots \ \cdot \underbrace{\begin{pmatrix}
    0 & b_{n} \\
    1 & a_{n} 
    \end{pmatrix}}_{M_X(n)}.
\end{equation}
%Eq. \ref{eq:pcf_to_mat_mult} should be interpreted graphically as moving along a horizontal trajectory in the following sense. 
The matrix $V(n)$ is understood as the ``state matrix'' of the system at the integer point $n=1,2,3,\dots$
The matrix $M_X(n)$ will represent the ``work matrix'' required for a one-step displacement from $n$ to $n+1$, changing states from $V(n)$ to $V(n+1)$. Eq. \ref{eq:pcf_to_mat_mult} then  describes the displacement from the integer point $1$ to $n+1$. At the limit $n\rightarrow\infty$, both ratios $p_{n-1}/q_{n-1}$ and $p_n/q_n$ of the column vectors of the matrix $V(n+1)$ tend to the limit of the continued fraction. %\vspace{3pt}

Consider now the infinite family of formulas, indexed by the integer parameter $\alpha$, listed in the table of Fig. \ref{fig:lattice_motivation}. Each formula converges to a fractional linear transformation of $\zeta(3)$. Adding the dependency on the index $\alpha$ to Eq. \ref{eq:pcf_to_mat_mult} for each individual formula gives rise to the equation:\vspace{-2pt}
\begin{equation} \label{eq:general_mat_mult}
    V(n+1,\alpha) = V(1,\alpha) \cdot M_X(1,\alpha) \cdot \ \cdots \ \cdot M_X(n,\alpha), 
\end{equation}
where the matrix $M_X(x,y)$ is defined as
\begin{equation} \label{eq:zeta3_M_X}
M_X(x,y)=\begin{pmatrix}
    0       & -x^6 \\
    1 & x^3 + (x+1)^3 + 2y(y-1)(2x+1) 
    \end{pmatrix}.
\end{equation}
Eq. \ref{eq:general_mat_mult} can be interpreted graphically as consequent displacements rightward along horizontal trajectories. 
We can arrange the horizontal trajectories corresponding to $\alpha = 1,2,3,\dots$ as parallel lines ordered from bottom to top. This arrangement completes a \emph{2-dimensional grid}. 
%by placing \emph{vertical} trajectories at every integer value $n=1,2,3,\dots$. 
For convenience, we relabel the parameters $(n,\alpha)$ by $(x,y)$. 

\begin{figure}[H]
    \vspace{-1cm}
    \centering
    \includegraphics[width=13.5cm]{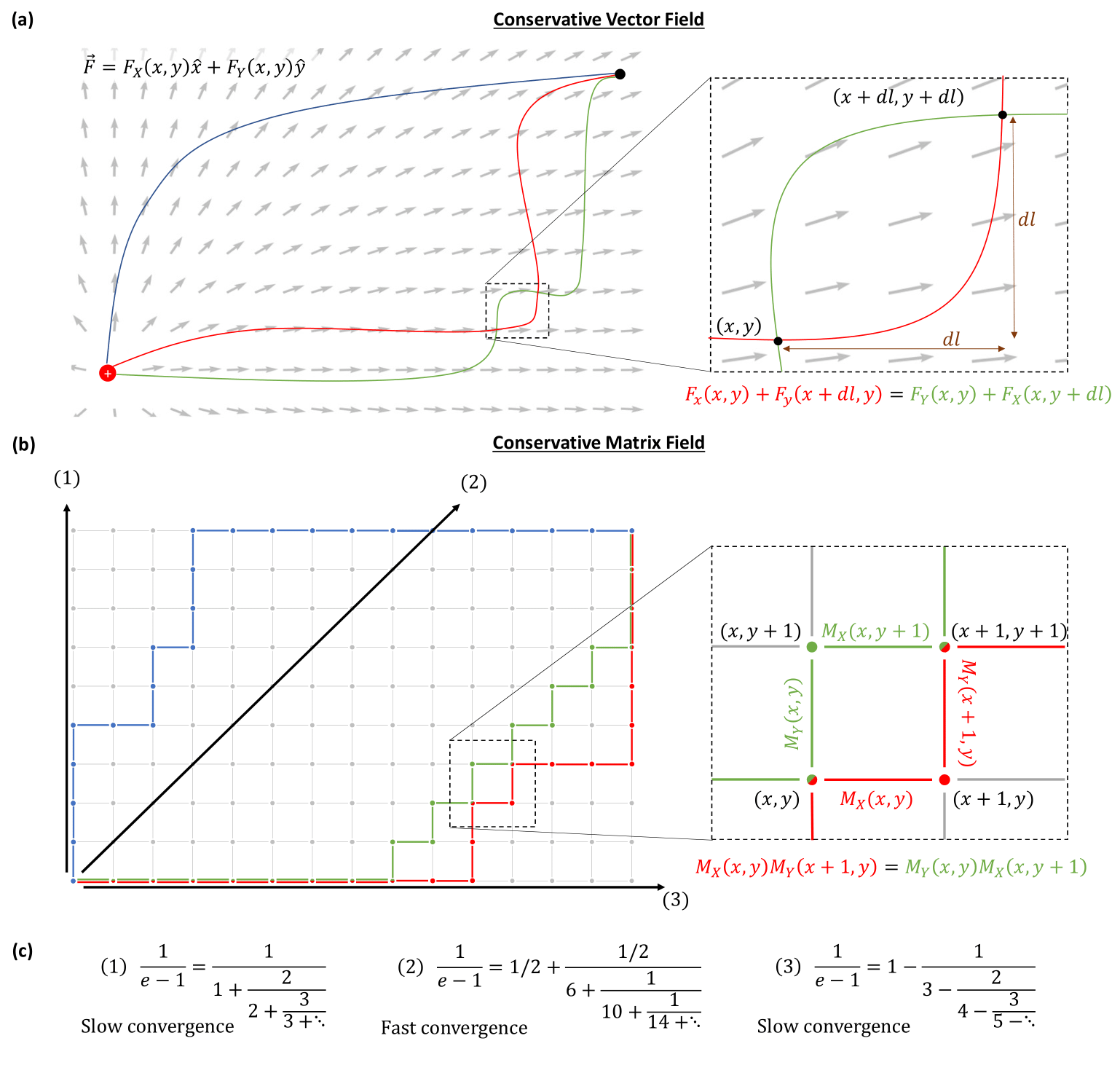}
    \vspace{-0.4cm}
    \caption{\textbf{Comparison of conservative vector fields and conservative matrix fields}.
    (a) Conservative vector fields exhibit path-independence, meaning that integrals taken over different paths yield the same result. 
    (b) Analogously, conservative matrix fields also offer path-independence, but for multiplications along the paths. 
    The resulting matrices only depend on the initial and final positions.
    Interestingly, conservative matrix fields are found to exhibit additional features: paths going to infinity represent continued fraction formulas. 
    (c) Each presented path provides a different formula of the same mathematical constant. The example continued fractions here are derived from the conservative matrix field of $e$ in Appendix D.1. %\ref{app:ssec:e_cmf}. 
    Such continued fractions can be derived for any constant that has a conservative matrix field. 
    We hypothesize that all continued fractions and infinite sums converging to a certain constant can be derived from a single conservative matrix field of that constant.}
    \label{fig:lattice_example}
    \vspace{-0.3cm}
\end{figure}

We now wish to examine the work necessary for a \textit{vertical} unit displacement in the grid, from $V(x,y)$ to $V(x,y+1)$. We denote the work matrix for such a displacement by $M_Y(x,y)$.
For this notation to be well-defined, the horizontal and vertical displacements must \emph{commute} in the sense of satisfying
\begin{equation} \label{eq:cl_conservative_condition} 
M_X(x,y) \cdot M_Y(x+1,y) = M_Y(x,y) \cdot M_X(x,y+1)
\end{equation}
for every pair of integers $x,y\ge1$. We illustrate this requirement in Fig. \ref{fig:lattice_example}(b).
In general, given an arbitrary $M_X(x,y)$, there is \textit{a priori} no reason why the connecting matrices $M_Y(x,y)$ should have a polynomial dependency on $(x,y)$. Surprisingly, in our example, $M_Y(x,y)$ takes the following form:
\begin{equation} \label{eq:zeta3_M_Y}
M_Y(x,y) = 
    \begin{pmatrix}
    -(x-y)(x^2-xy+y^2) & -x^6 \\
    1                & (x+y)(x^2+xy+y^2)
    \end{pmatrix}.
\end{equation}
The existence of polynomial matrices $M_Y(x,y)$ satisfying this equation in this example is quite remarkable.
%\vspace{3pt}

A \textbf{conservative matrix field} is defined as a choice of $2 \times 2$ matrices $M_X(x,y)$ and $M_Y(x,y)$ with integer polynomials in $x$ and $y$, such that the condition in Eq. \ref{eq:cl_conservative_condition} holds for a regime of $x,y$ in which the determinants of $M_X,M_Y$ are non-zero.
We call Eq. \ref{eq:cl_conservative_condition} the \textit{conservative property} (or the \textit{cocycle equation}). 
The conservative property guarantees that the work for a displacement along any trajectory in the 2-dimensional grid is given by a matrix that only depends on its initial and final coordinates. This path-invariance feature is analogous to the inherent property of conservative vector fields, with path integration replaced by matrix multiplications. Fig. \ref{fig:lattice_example} displays this striking analogy.

Drawing on this analogy, the state $V(x,y)$ of the conservative matrix field can be understood as a ``matrix potential'',
expressing the work matrix of any trajectory from a fixed origin (e.g., $(1,1)$) to the point $(x,y)$. Concretely, $V(x,y)$ can be calculated as a successive multiplication of matrices $M_X$ and $M_Y$ along any trajectory connecting the origin to the point $(x,y)$.
%The only limitation on the trajectories is to avoid any displacement by a matrix $M_X$ or $M_Y$ with determinant zero (which amounts to stepping out of the domain of definition). 
For example, two possible trajectories are
\begin{align*} 
    V(x,y) &=
    M_X(1,1)\cdot M_X(2,1) \cdot \ \cdots \ \cdot M_X(x-1,1) \cdot M_Y(x,1)\cdot M_Y(x,2) \cdot \ \cdots \ \cdot M_Y(x,y-1) \\[1pt]
    V(x,y) &= 
    \, M_Y(1,1)\cdot M_Y(1,2) \cdot \ \cdots \ \cdot M_Y(1,y-1) \cdot M_X(1,y)\cdot M_X(2,y) \cdot \ \cdots \ \cdot M_X(x-1,y)
\end{align*}
i.e., $V(x,y)$ can be computed either by first performing a rightward displacement of $x-1$ units and then an upward displacement of $y-1$ units, or vice versa.

We define the limit along a trajectory starting at the origin as the ratio of the entries in the right column of the potential matrix $V$ as we proceed on the trajectory toward infinity. In the special case of a horizontal trajectory in the example matrix field of Eqs. \ref{eq:zeta3_M_X},\ref{eq:zeta3_M_Y}, this limit corresponds to the continued fraction in the first row of Fig. \ref{fig:lattice_motivation}.
It is a remarkable observation that the limit along \textit{any} trajectory of our example conservative matrix field converges to $1/\zeta(3)$. This observation is proven in the complementary article \cite{CMF}. 
Building on this remarkable property, we can derive an abundance of new formulas of $\zeta(3)$, each from a different trajectory direction in the conservative matrix field. These formulas serve as the first outcomes from exploring the implication of the concept of conservative matrix fields.

It is even more remarkable that the property of a trajectory-invariant limit prevails across multiple conservative matrix fields, spawning infinite families of formulas for each mathematical constant, each tied to a distinct trajectory.

We produced a rich collection of conservative matrix fields (summarized in Appendix D) %\ref{app:all_cls}) 
that were all originally discovered via the Distributed Factorial Reduction algorithm, originating from infinite families of formulas analogous to that in Fig. \ref{fig:lattice_motivation}.
These matrix fields in turn led us to discover 
an analytic approach to constructing additional conservative matrix fields purely from the conservation property of Eq. \ref{eq:cl_conservative_condition} (summarized in Appendix H). %\ref{app:CMF_families}). 
This analytical construction shows that conservative matrix fields generalize the notion of rational Wilf--Zeilberger pairs \cite{A_eq_B_Zeilberger}.
Interestingly, our constructed matrix fields for constants such as $\pi$ and $\zeta(2)$ also demonstrated the property of having the same limit along all infinite trajectories, just as the matrix fields discovered by the Distributed Factorial Reduction algorithm. The occurrence of this peculiar phenomenon suggests the existence of a fundamental principle underlying a distinguished class of conservative matrix fields with well-defined limits.
The axiomatization of this class could broaden the applications of conservative matrix fields, from refining methods for numerical approximations and aiding in symbolic computations, to providing new paths for irrationality proofs.

The definition of a conservative matrix field requires two matrices $M_X(x,y)$ and $M_Y(x,y)$ that commute in the sense of Eq. \ref{eq:cl_conservative_condition}. This definition can be naturally generalized to encompass a set of $d$ pairwise commuting matrices
\[
    M_{X_1}(x_1,\ldots,x_d), \ \ldots, \ M_{X_d}(x_1,\ldots,x_d).
\]
We call such a structure a \emph{$d$-dimensional} conservative matrix field (each matrix is still $2\times2$). Appendix E %\ref{app:multidimensional_cmf} 
presents an example of a 4-dimensional conservative matrix field that converges to $\zeta(2)$ along every infinite trajectory. 

\subsection{Connections between constants via conservative matrix fields} \label{ssec:rational_shifts}
So far we have seen that each conservative matrix field generates infinite
different formulas of a single constant by following different straight trajectories. 
Each such formula can be directly translated to a continued fraction form \cite{ESMA}, and interestingly, the property of factorial reduction appears to be preserved in all of them (once a single direction had it).
These observations suggest an appealing hypothesis --- that all continued fraction formulas of a certain constant can be derived from different trajectories in a single conservative matrix field (of two or more dimensions). 

Further computational experiments reveal that conservative matrix fields can establish new relationships between various mathematical constants.
Specifically, by initializing the trajectory at a point other than $(1,1)$---for example, starting at $(1/2,1)$ (with the same $M_X$ and $M_Y$)---produces a sequence that converges to a different limit than the one starting at $(1,1)$, but arises from the same conservative matrix field. If this shift is integer, then the resulting limit will be related to the same constant by a linear fractional transformation. Conversely, when the shift is a non-integer rational number, in most cases the trajectory produces a formula that converges to a different limit that is not related to the original constant. 
For example, $\pi$ and $\sqrt{2}$ both emerge from the same conservative matrix field via a shift of $1/2$ of the trajectory initialization along the $y$ axis. A similar connection exists for other constants such as $\zeta(2)$ and Catalan's constant, or $\zeta(3)$ and $\pi^3$.
The conservative matrix field can also connect $\pi$ and $\ln(2)$, by re-scaling the $x$ axis $x \to x/2$, and following a trajectory along the $x$ direction. 

These connections transfer formulas that were developed for one constant to the other constant, and may help translate other properties between the constants. e.g. the 4-dimensional conservative matrix field of $\zeta(2)$ in Appendix E %\ref{app:multidimensional_cmf} 
directly gives rise to a similar matrix field for $G$. 
Constants that share their matrix field up to such rational shifts can be placed at the same level of the hierarchy --- the matrix field guarantees that they are derived from recurrence formulas with the same structure, with polynomials of the same degree. 

Different trajectories in a conservative matrix field can express the same constant by formulas with vastly different attributes. Altering the trajectory may enable generating formulas with desired attributes like more efficient convergence, crucial for proving the irrationality of various constants.
In the section below, we follow this observation and analyze straight trajectories with different slopes in the 2-dimensional grid. 
We show how to provide quantitative formulas that yield different lower bounds on the irrationality measure of the target constant.

\subsection{Proofs of irrationality}\label{ssec:irrationality_from_cmfs}
Ever since Ap\'ery's breakthrough proof of the irrationality of $\zeta(3)$, there have been various attempts to generalize his argument to provide irrationality proofs for other constants. In this section, we will show that the conservative matrix field of $\zeta(3)$ can be used to provide a systematic approach to prove its irrationality. The same approach
extends to different conservative matrix fields, opening a pathway for irrationality proofs for other constants. 

All these proofs are based on the premise that 
any continued fraction that converges to a constant produces a sequence of rational approximations, whose fast enough convergence implies the irrationality of the constant. This argument can be quantified by
the Liouville--Roth irrationality measure \cite{intro_num_theo_ir_measure}. The irrationality measure of a real number $L$ is the largest possible number $\delta$ for which there exists a sequence of distinct rational numbers $\{p_n/q_n\}$ that converges to $L$ and satisfies the inequality
\begin{equation} \label{eq:dioph}
\left|\frac{p_n}{q_n}-L\right| < \frac{1}{q_n^{1+\delta}}
\end{equation}
for sufficiently large $n$. Rational numbers have irrationality measure $\delta=0$ and irrational numbers have irrationality measure $\delta\geq1$
by a classical result of Dirichlet. 
These two facts combined give rise to the next irrationality criterion: if an infinite sequence of rational numbers converging to $L$ has a positive irrationality measure, then $L$ is irrational. 

It is possible to define the irrationality measure $\delta$ \textit{of a rational sequence} as the largest value that satisfies Eq. \ref{eq:dioph} for a sub-sequence. This $\delta$ bounds the irrationality measure of the constant to which the sequence converges, and can thus be used to prove the irrationality of the constant. For example, Apéry's continued fraction in Eq. \ref{eq:apery_pcf} has an irrationality measure $\delta\approx 0.08$, which is how Apéry proved the irrationality of $\zeta(3)$ \cite{apery,poorten}.

Conservative matrix fields enable deriving a closed-form formula for $\delta$ that generalizes the work by Apéry to arbitrary polynomial continued fractions, providing a parametric family of rational approximations for each constant.
The formula followed from our observation of \textit{a strong form of factorial reduction} that appears in matrix fields (after balancing their degrees \cite{endnote_balanced_matrix}):
Let $g(x,y)=\textrm{GCD}(V(x,y))$ be the greatest common divisor calculated over the four entries of the matrix $V$. 
Numerical experiments show that most of the conservative matrix fields that we found \cite{endnote_fr_in_e_cmf}
have their reduced quotients $V_{ij}(x,y)/g(x,y)$ all grow exponentially rather than factorially, for each straight trajectory $t(n)=(x(n),y(n))$ that goes to infinity in the 2-dimensional grid.
We hypothesize that this remarkable strong factorial reduction property is intrinsic to all conservative matrix fields in which the polynomials in the four quotients have the same degree.
The base of the exponential growth $s$ is trajectory dependent, and can be derived from any of the matrix entries $ij$, as $s^n \approx V_{ij}(t(n))/g(t(n))$.

Our closed-form $\delta$ formula thus provides a different value for each trajectory: 
\begin{equation} \label{eq:delta_from_cl_traj}
1 + \delta = \frac{ \ln{|e_{\max}/e_{\min}|}}
                { \ln{s} },
\end{equation}
where $|e_{\min}| \le |e_{\max}|$ are the eigenvalues of the matrix that represents a step along the trajectory $t(n)=(x(n),y(n))$ \cite{endnote_matrix_trajectory} in the limit of $x,y$ going to infinity.

\subsection{The irrationality of $\zeta(3)$} \label{ssec:zeta3_irrationality}
\begin{figure}[H]
    \centering
    \includegraphics[width=16cm]{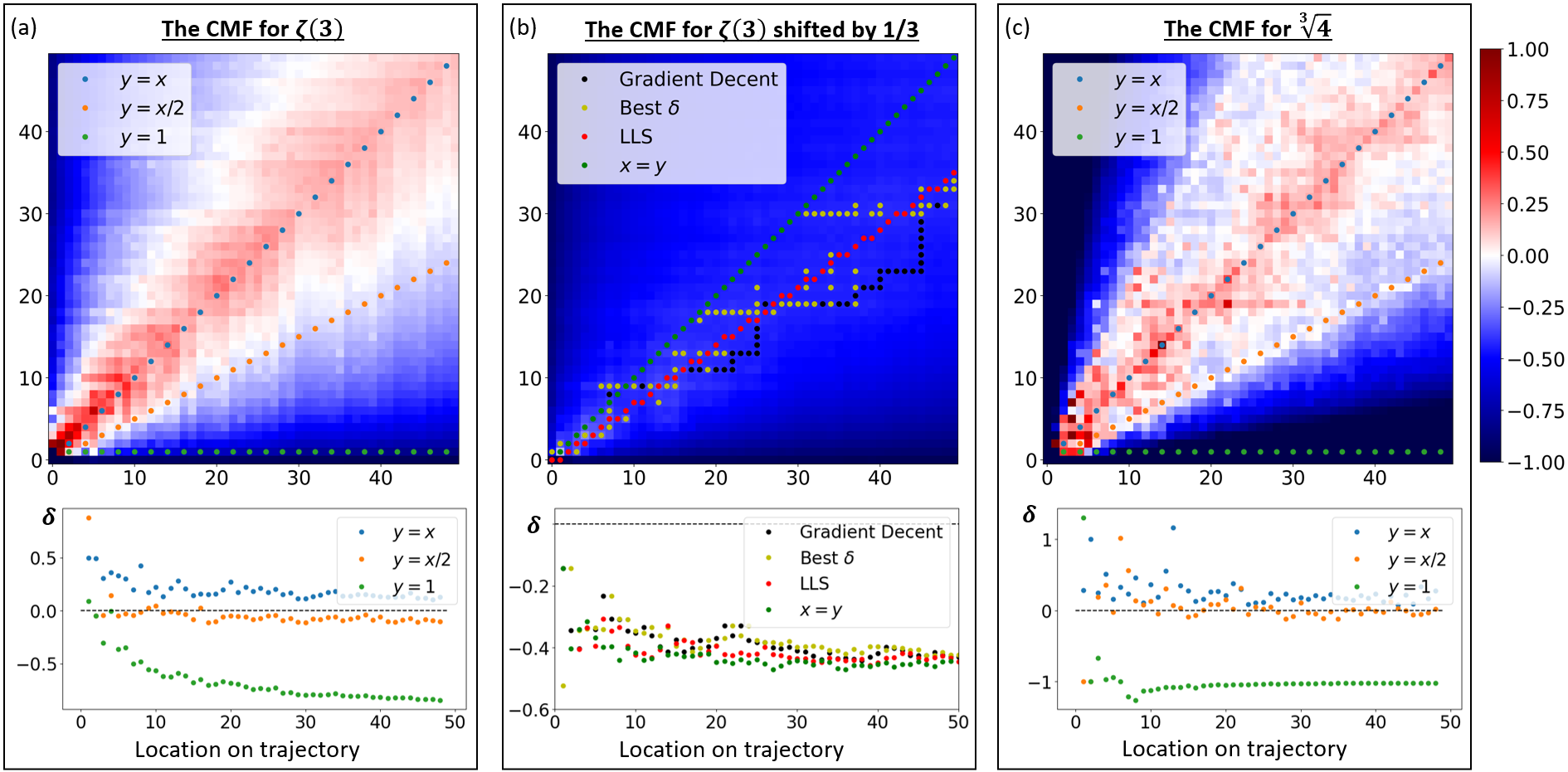}
    \caption{\textbf{Using conservative matrix fields to extract irrationality measures}.
    (a) The extracted irrationality measures from the conservative matrix field of $\zeta(3)$. Every infinite trajectory that remains in the red area yields a sequence with $\delta>0$, which can prove the irrationality of $\zeta(3)$. The optimal $\delta$ is along the $x=y$ trajectory, which is equivalent to Apery's continued fraction \cite{apery,poorten,CMF}.
    (b) The conservative matrix field of $\zeta(3)$ shifted by $1/3$ along the $x$ axis. The result is a conservative matrix field that converges to a different constant: one that combines $\zeta(3)$ and $\pi^3$; it is not known whether this constant is rational or irrational. The optimal $\delta$ is no longer along the $x=y$ trajectory. We present two methods to detect the optimal $\delta$ trajectory through optimization algorithms, depicted in Appendix F.1 %\ref{app:ssec:cmf_traj_algos} 
    and marked here by colored dotted trajectories. Although this conservative matrix field does not contain a trajectory with $\delta>0$, the same methods provide $\delta>0$ for other constants (Appendix I), %\ref{app:ir_meas_cmf_family}), 
    and may yet provide a positive $\delta$ for this constant in a higher dimensional extension of this matrix field.
    (c) The extracted $\delta$ for a conservative matrix field corresponding to $\sqrt[3]{4}$. The optimal trajectory is $x=y$, offering an irrationality-proving $\delta>0$.}
    \label{fig:zeta3_delta}
    \vspace{-0.4cm}
\end{figure}

To find the trajectory that offers the best $\delta$ in a conservative matrix field, we extract a numerical value $\delta=\delta(x,y)$ 
at every point in the 2-dimensional grid. 
$\delta(x,y)$ can be extracted from the following equation, which arises from imposing equality in \eqref{eq:dioph}:
\begin{equation} \label{eq:dioph_CMF}
\left|\frac{\Tilde{V}_{12}(x,y)}{\Tilde{V}_{22}(x,y)} - L\right| = \frac{1}{\Tilde{V}_{22}(x,y)^{1+\delta}}.
\end{equation}
Here, $\Tilde{V}(x,y) = V(x,y)/g(x,y)$ is the \textit{reduced} potential for every location $(x,y)$ in the matrix field.
Doing so produces a function $\delta:\!\mathbb{N}^2 \rightarrow \mathbb{R}$ on the 2-dimensional grid. The limit of this function at infinity along each direction provides the value of the irrationality measure of the rational approximation corresponding to that direction. We thus search for the direction at which this limit is maximal.
The resulting $\delta(x,y)$ for a few examples are presented in Fig. \ref{fig:zeta3_delta}. 
Every location where $\delta>0$ is colored in red. Hence, assembling a sequence along a trajectory contained in the red regime of the figure yields a sequence that demonstrates the irrationality of the constant. 

To find an exact expression for the trajectory $x=y$, we define $M_\text{traj}(n) = M_X(n,n) \cdot M_Y(n+1,n)$.
The resulting matrix (after balancing the degree \cite{endnote_balanced_matrix}) is
\begin{equation} \label{eq:z3_best_traj}
M_\text{traj}(n) = n^3 \cdot 
\begin{pmatrix}
   -(n+1)^3        &     -6n^3 - 9n^2 - 5n - 1  \\
   6(n+1)^3        &     35n^3 + 54n^2 + 30n + 6
   \end{pmatrix},
\end{equation}
for which the eigenvalues are (for $n\rightarrow\infty$) $e_{\pm}=17 \pm 12\sqrt{2}$. 
A direct numerical test shows that $\log s \approx 6.5$, 
and thus $\delta \approx 0.08$, matching
the irrationality measure of the continued fraction found by Apéry \cite{apery,poorten}, which was used in the first proof of the irrationality of $\zeta(3)$. Our complementary work \cite{CMF} shows how to use this conservative matrix field to prove the irrationality of $\zeta(3)$. In Appendix F.2, %\ref{app:ssec:pcf_to_cmf_equivalence}, 
we show the equivalence between this matrix representation and Apéry's continued fraction.

Consequently, the conservative matrix field can be seen as an underlying algebraic structure that lies behind Apéry's continued fraction and constitutes his irrationality proof. This finding further motivates searching for similar conservative matrix fields for other values of the Riemann zeta function and other mathematical constants. As an example, the same method, applied on a conservative matrix field for $\zeta(2)$ (Appendix D) %\ref{app:all_cls}) 
also provides a way to prove the irrationality of this constant, providing $\delta \approx 0.09$.
Generalizing beyond $\zeta(2)$ and $\zeta(3)$,
it remains to be discovered whether similar matrix fields exist for other $\zeta(n)$,
and whether they could help prove the irrationality of these constants.
Appendix G %\ref{app:ZZZ_lattice} 
shows a similar method applied for higher zeta values, yielding sequences with non-trivial irrationality measures, though not positive values so far.

\subsection{Results for $\zeta(5)$ and the quest to prove its irrationality}
It would be of great interest to create a conservative matrix field for $\zeta(5)$, for which irrationality is a long-standing open question \cite{Zudilin_z5_ir_measure,zudilin_one_zetas_is_ir}. Although we have not yet found such a matrix field, this section describes several formulas for $\zeta(5)$ (Table \ref{table:zeta5_res}) that may serve as the foundation for constructing a conservative matrix field.
Part of the formulas in Table \ref{table:zeta5_res} combine several values of the Riemann zeta function, hinting that they belong to different conservative matrix fields (or to different rational shifts of the same matrix field).
Three of the discovered formulas in the table did not fit any combination of zetas (in numerical tests using PSLQ) and are thus presented with their numerical values.

\begin{table}[H]
\begin{center}
\def\arraystretch{1.5}
\begin{tabular}{ |c|c|c| } 
\hline
    $a_n$   &   $b_n$   & Formula \\
\hline
    $n^5 + (n+1)^5 + 6(n^3 + (n+1)^3)                   $     &   $-n^{10}$      &   $\frac{2}{2\zeta(5)+6\zeta(3)-9}$ \\ 
\hline
    $n^5 + (n+1)^5 + 6(n^3 + (n+1)^3) - 4(2n+1)         $     &   $-n^{10}$      &   $\frac{2}{2\zeta(5)-2\zeta(3)+1}$ \\ 
\hline
    $n^5 + (n+1)^5 + 16(n^3 + (n+1)^3) - 4(2n+1)         $     &   $-n^{10}$      &   $\frac{64}{64\zeta(5)+176\zeta(3)-273}$ \\ 
\hline
    $8 (n^5 + (n+1)^5) - 15(n^3 + (n+1)^3) + 9(2n+1)$     &   $-64n^{10}$      &  $1.20426...$ \\ 
\hline
    $8 (n^5 + (n+1)^5) - 12(n^3 + (n+1)^3) + 7(2n+1)         $ &
    $-64n^{10}$  &
    $2.45174...$
    \\
\hline  
    $8 (n^5 + (n+1)^5) + 20(n^3 + (n+1)^3) - 5(2n+1)         $ &
    $-64n^{10}$  &
    $22.8410...$
    \\
\hline

\end{tabular}
\end{center}
\vspace{-0.5cm}
\caption{\textbf{Sporadic results
suspected as related to $\zeta(5)$}, 
found via the Distributed Factorial Reduction algorithm for polynomial continued fractions of degree 5. Part of the formulas were linked to a combination of $\zeta(5)$ and $\zeta(3)$.
}
\label{table:zeta5_res}
\end{table}

The Distributed Factorial Reduction algorithm discovered infinite families of formulas encompassing different combinations of zeta function values, such as Eq. \ref{eq:zzz_general}. Although we did not identify a matrix field that generalizes these continued fractions, we propose a method that uses them to extract meaningful irrationality measures. We demonstrate this method on $\zeta(5)$.
We construct new sequences that converge to $\zeta(5)$ by assembling linear combinations of continued fractions. 
To find the coefficients of the combinations $c_i$, we use the closed-form expression of the limit of the continued fraction from Eq. \ref{eq:zzz_general}, denoted by $\hat\zeta(s,R)$. 
For each set of parameters $R_i$, we can determine rational coefficients $c_i$ such that:
\begin{equation} \label{eq:zzz_lattice_sum_of_pcfs}
%\zeta(5) = \frac{c_2}{\hat\zeta(s=2,R_2)} + \frac{c_3}{\hat\zeta(s=3,R_3)} + \frac{c_4}{\hat\zeta(s=4,R_4)} +\frac{c_5}{\hat\zeta(s=5,R_5)}
\zeta(5) = c_2\cdot\hat\zeta(s=2,R_2) + c_3\cdot\hat\zeta(s=3,R_3) + c_4\cdot\hat\zeta(s=4,R_4) +c_5\cdot\hat\zeta(s=5,R_5)
%\zeta(5) = c_2\hat\zeta(s=2,R_2) + c_3\hat\zeta(s=3,R_3) + c_4\hat\zeta(s=4,R_4) +c_5\hat\zeta(s=5,R_5)
\end{equation}
To construct a sequence of rational numbers that converge to $\zeta(5)$, we substitute each $\hat\zeta(s,R)$ by the sequence of convergents of its continued fraction form (Eq. \ref{eq:zzz_general}).
%Substituting each rational convergent $\hat\zeta(s,R)$ by successive convergents of its continued fraction (Eq. \ref{eq:zzz_general}) yields a sequence of rational numbers that converges to $\zeta(5)$.
% Substituting the corresponding continued fractions (Eq. \ref{eq:zzz_general}) in place of each $\hat\zeta(s,R)$ yields a new sequence that converges to $\zeta(5)$.
%where $\hat\zeta(s,R)$ is the ZigZagZeta continued fraction from Eq. \ref{eq:zzz_general} with degree $s$ and a root shifted by $R$. 
Varying the $R_i$ parameters, we can generate infinitely many sequences for $\zeta(5)$, organized in a structure reminiscent of Fig. \ref{fig:lattice_motivation}. %, similar to a conservative matrix field.
%Sampling successive terms from consequent sequences (of different $R_i$ values) creates new sequences of faster convergence, similar to Fig. \ref{fig:lattice_motivation}.
The analysis of the convergence rates for different $R_i$ values yields nontrivial irrationality measures for $\zeta(5)$ (Appendix G). %\ref{app:ZZZ_lattice}). 
This method may help improve the irrationality measure beyond the recently achieved record value \cite{Zudilin_z5_ir_measure}.

% \begin{equation} \label{eq:zzz_deg5}
% \frac{1}{\zeta(5)-\zeta(4)+\zeta(3)-\zeta(2)+1}=2+\cfrac{-2}{49+\cfrac{-1536}{356+\cfrac{-78732}{\ddots + \cfrac{-(n+1)n^9}{n^5 + (n+1)^5 + (n+1)^4 + \ddots}}}}.
% \end{equation}

\section{Discussion} \label{sec:discussion}
The incorporation of the distributed algorithm caused a dramatic increase in the number of formula candidates obtained by the Ramanujan Machine,
which, in turn, posed new kinds of algorithmic challenges. Specifically, we are in need of an automated method to \textit{identify} the relevant constants that connect to each formula candidate, as well as a method to \textit{generalize} a set of formulas to parametric families and to conservative matrix fields.
These challenges are discussed in the next subsections.

\subsection{Algorithms for finding connections between continued fractions and mathematical constants} \label{ssec:PSLQ}
The Distributed Factorial Reduction algorithm identifies promising candidate continued fractions but does not identify the mathematical constants to which they converge.
To find a closed-form formula for the continued fraction, we try to match a selection of candidate constants to the numerical value of the discovered continued fraction using the integer-relation algorithm PSLQ \cite{PSLQ1992,PSLQ1999}. 
Appendix A % \ref{app:trm_desc} 
compares this approach to other algorithms that we had previously developed \cite{TRM} for this purpose.

The PSLQ algorithm accepts a vector of real numbers $z_i$, and yields a vector of integers $c_i$ such that $\sum{c_iz_i} = 0$ up to a preset error.
In our simplest use of PSLQ, for a continued fraction that evaluates to a numerical value $v$ and a candidate constant $\eta$, the input vector is $(1, \eta, -v, -v\eta)$. 
If the algorithm finds a set of integers ${c_0,c_1,c_2,c_3}$, then the formula
$\frac{c_0+c_1\eta}{c_2+c_3\eta} = v$ holds.
Most of the results found in this work were based on this application of PSLQ, resulting in conjectured formulas such as Eq. \ref{eq:zeta2_family_member}. 

Importantly, the input vector of PSLQ can be extended to include several constants and powers thereof, expanding the scope of our findings to more complex formulas. 
Successful examples of operating PSLQ with such extended vectors led to the most intriguing results of this work, including the conjectural formula of Eqs. \ref{eq:mixed_z3_z5} and \ref{eq:zzz_general} (more information in Appendix C). % \ref{app:all_pcfs}).
Our use of PSLQ can be taken further to include other functions of constants (e.g. square roots, exponents) and even to connect several continued fractions.
We hypothesize that eventually, a refined version of PSLQ will find a closed-form formula for all the continued fractions that were found in this work.

\subsection{Result verification} \label{ssec:results_verification}
Conjectures found by data-driven algorithms in experimental mathematics are tested via their numerical approximations and are hence bound to the possibility of being false positives 
(until they are proven \cite{dougherty2020automatic,note_on_rm}).
In order to reduce the occurrence of false detection, we calculate the continued fraction to a greater depth \cite{endnote_dynamic_dept_for_pcf} and compare it with additional digits of the constants. 
Most formulas reported in this work have been tested to at least 100 digits of precision.
However, the slow convergence rate of certain formulas (e.g., the first few members of the $\zeta(3)$ infinite family in Fig. \ref{fig:lattice_motivation}) makes the task of guaranteeing that precision in these cases computationally infeasible. 

To acquire additional confidence in the slowly converging formulas, we are again inspired by experimental physics, where one can mitigate errors by identifying \textit{patterns} that support an underlying model. In such cases, multiple observations---each possibly with low precision---can together lead to the discovery of a model with high confidence, if a shared pattern supports that model.
The same process translates to our investigation in experimental mathematics.
We compensate for the low precision in slowly converging continued fractions by identifying shared patterns, like \textit{parametric families} that generalize several formulas (e.g., Fig. \ref{fig:lattice_motivation} and Appendix C). %\ref{app:all_pcfs}).
In such a case, even though every formula is verified only up to a limited precision, the appearance of a family of formulas provides more confidence in the validity of each of its members. For example, the first two members of the $\zeta(3)$ infinite family from Fig. \ref{fig:lattice_motivation} acquire less than 50 correct digits when calculated to a depth of $10^5$. Still, the faster convergence of other formulas in that family adds to our confidence in the slower converging members.

In physics, for additional verification, candidate models can be used to generate new predictions for targeted experiments in previously untested parameters. The successful validation of these predictions serves to enhance our confidence in the physics model.
Translating this approach to our case in experimental mathematics, we tested the validity of every candidate parametric family of formulas 
by substituting new (rational) parameters and evaluating the formula numerically.
%The validation of a new member formula in the family increases our confidence in the entire family and, thus, also in every specific member formula of limited precision.
Breaking the analogy with physics, it is not straightforward to assess the numerical error for particular continued fractions. We estimate the error based on the rate of convergence of the continued fraction (Appendix B). %\ref{app:result_verification}).

Lastly, we consider the innate beauty of mathematical representations as a tool for verification. 
%Mathematical constants are, among other things, a compact way to represent fundamental information.
%One can use infinitely many digits to describe $3.14159...$, yet, more efficiently, we may just use the single letter $\pi$ or a short recurrence formula like a continued fraction. 
%We use PSLQ to represent a finite decimal approximation of a continued fraction using several integers and mathematical constants.
PSLQ is more successful when finding a formula that holds to a large precision using only small integers and only a few constants. 
When the formula suggested by PSLQ is more compact, we usually consider it more ``beautiful'', perceiving the formula as more likely to be correct. 
A quantitative rule of thumb for success in PSLQ is searching for results in which the total number of digits in all the integers found by PSLQ is significantly smaller than the number of input digits (resulting from the numerical evaluation of a continued fraction).
%The same consideration also explains how PSLQ fails when it fits a formula with integers that are too large: PSLQ can fit \textit{any} decimal approximation to \textit{any} mathematical constant. For example, PSLQ can find a (wrong) relation between $e$ and $\pi$ that holds up to the first 5 digits using the integers 314157 and 100000 as: $\frac{314157+e}{100000} = \pi$. 
From the point of view of the compactness of the representation, smaller integers imply that most of the information provided by the continued fraction is held by the constant rather than by the integers.
We also gain increased confidence in a new candidate result of PSLQ when the resulting integers remain unchanged after running PSLQ again with more digits as input.

\newpage
\subsection{Outlook and open questions}
The new algorithmic approach shown here is the most successful ever in finding conjectured formulas for mathematical constants. 
The sheer number of new formulas led to new challenges and motivated clustering of formulas of the same constant.
In certain cases, we identified such connections, which in turn unveiled a novel mathematical structure---the conservative matrix field. 
This structure has created infinite families of continued fractions, enabled irrationality proofs, and revealed a hidden hierarchy connecting different mathematical constants.

Our algorithm-assisted research has inspired numerous intriguing, open questions. 
Many of these questions concern the properties of conservative matrix fields. 
The foremost question is whether indeed a unique conservative matrix field exists for each constant, a premise that our findings appear to support so far.
If corroborated, this concept could have intriguing implications, suggesting that a unified matrix field may encapsulate all the formulas corresponding to each specific constant. 
We must also establish the maximum dimension (number of matrices) within the matrix field of each mathematical constant, as this attribute could become a fundamental characteristic of each constant.

Capturing additional constants may be possible by generalizing the concept of a conservative matrix field to larger matrices beyond $2\times2$, or to other analogous mathematical structures.
It is also important to find whether $2\times2$ conservative matrix fields exist at all for an arbitrary polynomial degree (Appendix H %\ref{app:CMF_families} 
shows matrix fields up to degree 3).

The hierarchy uncovered using conservative matrix fields reveals a classification of constants based on the complexity of the formulas producing them. This ordered classification offers a valuable avenue for exploring shared properties between constants of the same order. Consequently, conservative matrix fields emerge as a potent and promising tool for identifying invariants among mathematical constants, paving the way for novel insights into their fundamental nature.

At present, algorithmic-assisted research strategies are still at their infancy.
Our work has detailed the process of large-scale experimental exploration and discovery, unveiling many numerically-validated formulas, which we subsequently generalized into a unified mathematical structure.
Traditionally, such generalizations are attributed to moments of inspiration---when a mathematician discerns a pattern or creates a novel structure. 
We foresee this process of generalization becoming algorithmically-assisted in the future. 
This can be achieved through techniques such as feature extraction and pattern recognition applied to the wealth of data generated by large-scale mathematical experiments.
Such algorithms could generate high-quality hypotheses, automatically proposing generalizations for researchers to review, validate, and ultimately prove.

We could consider the heuristic of factorial reduction as an intriguing case study of automating the act of generalization and hypothesis. This heuristic, which led to most of the results presented here, was discovered by (manually) noticing an emergent pattern. In future works, identifying statistical anomalies or surprising order within experimental results (like the unexpected prevalence of factorial reduction in formulas of constants) could be done algorithmically as well---potentially identifying effects and structures that may otherwise be overlooked.

Looking ahead, algorithmic approaches in experimental mathematics are set to provide ever-stronger methods to study long-standing open questions. In the coming years, more sophisticated algorithms will be tailored to generate conjectures of growing complexity. Such conjectures, formulated automatically, could 
provide new clues and accelerate progress across various fields of mathematics, tackling profound questions such as the structure and properties of fundamental constants.

\subsection{Acknowledgements}
This research is supported by the generosity of Eric and Wendy Schmidt by recommendation of the Schmidt Futures Polymaths program.
We are grateful to the volunteers in the BOINC community whose contribution made this discovery possible.

\newpage
\bibliographystyle{unsrt} 
\bibliography{bibfile}

\begin{thebibliography}{10}
\bibitem{endnote_formula_is_pcf}
By default, we shall use the term ``formula'' to refer to the polynomial
  continued fraction formulas, such as \eqref{eq:apery_pcf}. Our approach
  covers other types of formulas as well.
\bibitem{endnote_EulerCF}
We can test an infinite sum for the factorial reduction property by converting
  it to a continued fraction using Euler's continued fraction formula, or by a
  direct examination of the growth of the reduced numerators and denominators
  of the partial sums.
\bibitem{endnote_balanced_matrix}
We transform the matrices to reduce their maximal degree. This transform is
  motivated by the computational effort of matrix multiplication, which is
  dominated by the largest element in the matrix. For example, the matrix in
  equation \ref{eq:zeta3_M_X} contains $a_n \propto n^3$ and $b_n \propto
  n^{6}$. By splitting the roots of $b_n$ to two groups, and multiplying one
  group to the preceding matrix, we decrease the largest matrix entry to be
  $O(n^{3})$ instead of $O(n^{6})$. This process offers a significant
  computational advantage, while only changing the produced expressions by a
  linear fractional transform.
\bibitem{endnote_fr_in_e_cmf}
Except for the conservative matrix field of $e$, where the convergence is also
  faster and thus a nontrivial $\delta$ is still extracted.
\bibitem{endnote_matrix_trajectory}
Let $M_\text{t}(n)$ be a matrix that represents the direction of the
  trajectory. For example, a diagonal going in $45^{\circ}$ is represented by
  $M_\text{t}(n) = M_Y(n,n) \cdot M_X(n,n+1)$, and a diagonal going in
  $30^{\circ}$ relative to the $x$ axis is represented by $M_\text{t}(n) =
  M_X(2n,n) \cdot M_X(2n+1,n) \cdot M_Y(2n+1,n+1)$.
\bibitem{endnote_dynamic_dept_for_pcf}
The number of terms is decided as the calculation takes place, depending on the
  rate of convergence, which determines the (increaseing) depths at which we
  sample the decimal approximations.
\end{thebibliography}

\newpage
\appendix
%\documentclass[main.tex]{subfiles}
%\begin{document}
%\appendix

\section{Past algorithms in the Ramanujan Machine project}  \label{app:trm_desc}
This section describes the algorithms for discovery of formulas for mathematical constants that were developed as part of the Ramanujan Machine project prior to this work. 
\begin{itemize}
    \item \textbf{Meet-in-the-Middle} \cite{TRM} - Suggests conjectured formulas in the form of polynomial continued fractions based on brute-force matching of decimal approximations. The algorithm computes polynomial continued fractions from a user-defined scheme to a predetermined depth, 
    creating a decimal approximation of the continued fraction. 
    The decimal approximation is then matched against a pre-calculated list of decimal approximations of rational functions of fundamental constants to identify potential formulas. 
    To ensure efficient execution, the algorithm incorporates bloom filters, caching of the sequences $a_n$ and $b_n$, and other optimizations. The following section provides a more comprehensive description of this algorithm. 
    \item \textbf{Descent and Repel} \cite{TRM} - Proposes conjectured formulas in the form of polynomial continued fractions (but applicable to arbitrary forms) by solving an optimization problem via an altered version of gradient descent combined with rounding to the nearest integer parameters. The gradient descent is executed with several points in parallel, incorporating a Coulomb-like repulsion between the points to enable a better cover of the search space.
    \item \textbf{Enumerated Signed-Continued-Fraction Massey Approve (ESMA)} \cite{ESMA} - Finds conjectured formulas in the form of signed continued fractions (where $b_n$ is a sequence made from $\pm 1$) by identifying patterns in integer sequences. For every sequence of $\pm$ signs, each mathematical constant has a unique signed continued fraction expansion in the form of an integer sequence. ESMA searches for patterns in such integer sequences using a modification of the Berlekamp--Massey algorithm.
\end{itemize}

All of these algorithms are limited by the requirement to specify the target constants. i.e., one must guess what constants are expected to be the exact convergence limit of the continued fractions.
The choice of the said constants is often based on intuition, or extracted from within a prescribed finite list. 
The Distributed Factorial Reduction algorithm bypasses this limitation, as it searches for continued fractions without relying on a convergence to any specific constant.
The following section provides a brief review of the Meet-in-the-Middle algorithm, exemplifying the above-mentioned challenge of choosing the target constant. The section then concludes by explaining how factorial reduction serves to bypass this challenge.

\subsection{The Meet-in-the-Middle algorithm}

This algorithm receives a predetermined constant $\eta$ for every execution and aims to find expressions of the form:
\begin{equation} \label{pcf_appendix}
\frac{a+b \cdot \eta}{c+d \cdot \eta} = 
\displaystyle
a_0 + \cfrac{b_1}
  {a_1 + \cfrac{b_2}
    {a_2 + \cfrac{b_3}
      {\ddots + \cfrac{b_n}
        {a_n + \ddots}}}}
\end{equation}
with $a,b,c,d \in \mathbb{Z}$. More general versions of the algorithm enumerate over more complex expressions in the left-hand-side. Either way, the algorithm has to enumerate over the spaces of values for each side of the equation separately and to compare them.

First, the algorithm calculates many left-hand-side expressions.
These expressions are stored in a hash table, and the key assigned to every entry is the first 10 digits of the decimal approximation. The keys are stored in a bloom filter, which allows for quick queries to determine whether an element is in the filter.

Next, the algorithm iterates over all polynomial continued fractions with numerators and denominators with fixed polynomial degrees, and with integer coefficients in a pre-specified interval. Each continued fraction is calculated to depth 30, and the first 10 digits of the resulting decimal approximation are computed. This value is then searched for in the bloom filter. If the value is found, the algorithm indicates a candidate result that is later tested further. Variants of the algorithm operate with a different number of stored digits and a different depth of calculation.

Finally, to eliminate false positives and identify the remaining formulas, the algorithm calculates the candidate polynomial continued fractions to higher precision by evaluating them to depth 10,000. The algorithm then compares the first 100 digits to the exact formula.

\subsection{Limitations of Meet-in-the-Middle that were alleviated in the Distributed Factorial Reduction algorithm}

While being very successful, Meet-in-the-Middle had several drawbacks that limited our ability to distribute it, and restricted the continued fractions that it can identify.

Firstly, the algorithm requires the storage of a bloom filter in memory and a hash table on disk. This not only restricts the number of potential expressions on the left-hand-side of Eq. \ref{pcf_appendix} that we can check, but also complicates the distribution process, requiring the download of additional data for every off-site worker. This limitation is not present in the Distributed Factorial Reduction algorithm, since it identifies continued fractions without requiring prescribing the constant in the left-hand-side.

Secondly, the algorithm assumes that the continued fraction converges quickly enough to stabilize its first few digits after a small number of iterations. Consequently, continued fractions that converge slowly will not be detected under this assumption. For example, the second formula in the infinite family of $\zeta(3)$ presented in Fig. \ref{fig:lattice_motivation} only gets 11 correct digits after calculating the first 100k terms of the continued fraction. Therefore, such a formula cannot be detected by this algorithm. 
% alpha=2 gets 11 correct digits, alpha=3 gets 51 and alpha=4 gets 69 (all after 100K iterations
In contrast, the Distributed Factorial Reduction algorithm does not rely on decimal approximations to detect potential conjectures, instead using the greatest common divisor of $p_n$ and $q_n$. Thus, the convergence rate of a continued fraction does not prevent its detection. i.e., this algorithm can identify slow and fast continued fractions equally well.

\section{Verification of results} \label{app:result_verification}
This section discusses the challenges of using finite numerical approximations when finding conjectures, and the mechanisms we use to address these challenges.
As mentioned in Sec. \ref{ssec:results_verification}, these approximations are prone to introduce errors to the process and consequently lead to false positive results. These false positives are eliminated by the mechanisms detailed below, allowing us to estimate the magnitude of errors and translate that error into a confidence measure in each result. 

\subsection{Approximation error in the evaluation of a continued fraction}
Our algorithms use decimal representations for comparing the limit of a polynomial continued fraction and the fundamental constant expression $L$. Hence, when discussing the precision of a continued fraction $\mathrm{PCF}_n$ at a finite depth $n$, we refer to the number of digits in the fraction's approximation that are identical to its limit. A precision of $K$ means an approximation error $\epsilon_n = |\mathrm{PCF}_n - L| < 10^{-K}$.

We observes in numerical experiments that $\epsilon_n$ monotonically decreases with increasing depth for continued fractions of the type used in this work.
This observation suggests that for every required precision $K$, there exists an $N$ such that for every $n>N$ the first $K$ digits of the continued fraction calculated to depth $n$ are identical to the first $K$ digits of the limit $L$:
\vspace{-0.1cm}
$$\lfloor \mathrm{PCF}_n \cdot 10^K \rfloor = \lfloor L \cdot 10^K \rfloor $$
\vspace{-0.1cm}
Note that for some edge cases, this equation will not be satisfied. For example, if the limit is an integer and the sequence approaches it from below, we can get $\sum_{i=1}^{n} {9\cdot 10^{-i}} \to 1$, while at no point the digits match.
However, since comparing digits is efficient computationally, and almost all limits of interest are non-integer, this method satisfies our needs.

To find how many digits we can trust as %we can ``count on'' as 
already accurate, our algorithm attempts to find the largest $K$ for a finite predetermined $n$. To achieve this, we calculate the decimal approximation of the continued fraction for two different depths, $n_1 < n_2$. Since the error monotonically decreases, the first digit where $\mathrm{PCF}_{n_1}$ and $\mathrm{PCF}_{n_2}$ do not match identifies an upper bound on the precision $K$ offered by $\mathrm{PCF}_{n_1}$. By a careful choice of $n_1$ and $n_2$, we can decrease the upper bound calculated for $K$ and get a better approximation for it. 

Our algorithm finds this upper bound of $K$ and halts the calculation when the upper bound exceeds the desired precision $K$ by a confidence margin, or when we reach the maximal allowed depth. In the latter case, we return the best attainable $K$ at this depth, allowing us to collect some useful data even for continued fractions with slow convergence.

The choice of $n_1$ and $n_2$ is done by a form of exponential backoff algorithm. Most of our continued fractions converge at least at a polynomial rate, with the slowest convergence rate thus being $\epsilon_n \sim 1/n$. Such a convergence rate implies that some continued fractions will present the same incorrect digits over a wide range of terms. This phenomenon makes it more challenging to distinguish digits that match the limit from those that do not, and as a result makes evaluating $K$ harder.
To obtain a good approximation of $K$, we want a significant difference between $\mathrm{PCF}_{n_1}$ and $\mathrm{PCF}_{n_2}$, 
%leading to fewer identical digits between $\mathrm{PCF}_{n_1}$ and $\mathrm{PCF}_{n_2}$ and therefore a tighter bound on $K$. 
since this provides higher confidence in the accuracy of the digits of $\mathrm{PCF}_{n_1}$ that are still equal to $\mathrm{PCF}_{n_2}$, and therefore a tighter bound on $K$.

We sample the continued fraction at depths $n_1,n_2,...$, choosing $n_{l+1} - n_l\propto l$, which matches the sampling rate to the convergence rate in the slowest case of $\epsilon_n \sim 1/n$. This sampling rate makes the calculated decimal approximation satisfies the requirement for substantial difference between subsequent $\mathrm{PCF}_{n_l}$, resulting in a tight bound of $K$.
As a result of this process, we attain a decimal approximation of the continued fraction's limit, along with the number of digits we refer to as ``correct''.

\subsection{Identifying over-fitted formulas discovered by PSLQ}
To match a linear fractional transformation (i.e., a Mob\"ius transform) of a constant to a decimal approximation, we utilize PSLQ to find integer coefficients $\vec{a}=(a_0,a_1,a_2,a_3)$ that satisfy the equation $\frac{a_0+a_1 \eta}{a_2 + a_3 \eta} = \mathrm{PCF}_n$ for a given constant $\eta$ up to a given precision. Using the method described in the previous section, we employ finite approximations of the continued fraction to evaluate the first $K$ correct digits of the limit. 
Given the approximation error of the continued fraction ($\sim 10^{-K}$), we require PSLQ to find $\vec{a}$ that satisfies the formula to the same accuracy. 

The structure of linear fractional transformations %  utilized by PSLQ 
is highly expressive. By using large enough coefficients $\vec{a}$, the algorithm can always find a formula that matches any constant to any continued fraction up to any finite precision.
We should thus distinguish between true results and false-positives. 
A true result implies that the formula will remain accurate when extended to more digits.
A false-positive result arises from over-fitting. For example, the formula $\frac{314157 + e}{100000} = \pi$ is correct up to $10^{-5}$, but is obviously false. 

To distinguish over-fitted results, we consider that 
a decimal approximation using $K$ digits can represent $10^K$ different numbers. 
This amount sets a limit on the size of the PSLQ coefficients $\vec{a}$.
If the total number of digits in the $\vec{a}$ found by PSLQ is also $K$, it indicates that the digits alone can already express $10^K$ different numbers. Thus, the match is most-likely a coincidence, indicating a false-positive.
Conversely, if the collective number of digits in $\vec{a}$ is significantly smaller then $K$, it means that the result is more likely true, as the probability for an accidental match is very small. 
Specifically, to identify false results, we use the difference between the number of digits given by the decimal approximation $K$ and the total number of digits $l$ in the vector $\vec{a}$ of PSLQ. A larger value of $K-l$ implies high confidence in the formula (with the chance of an accidental match scaling as $\sim 10^{l-k}$), while $K\approx l$ or $K<l$ lead to disqualifying the conjecture.

This condition can be explained intuitively as expressing the same information of $K$ digits using less data. 
Unlike the decimal approximation, which conveys information solely through its digits, the formula uses the mathematical constant for extra representation capability (added to the digits of $\vec{a}$ that define the transformation). 
The mathematical constant thus enables a more efficient representation of the information.
In other words, true results can be understood as a means of data compression of the numerically computed digits using the mathematical constant to provide a compact form. 
Consider for example the special case where the limit of the continued fraction is exactly the constant itself. Then, the linear fractional transformation does not need to provide any digits ($\vec{a}=(0,1,1,0)$), hence we say that the representation is maximally compressed - all the information is in the constant. The smaller the coefficients of the linear fractional transformation, the closer it is to maximal compression.

\section{Selected families of continued fractions found by the Distributed Factorial Reduction algorithm}  \label{app:all_pcfs}

In this section, we show a collection of infinite families of continued fractions discovered by the Distributed Factorial Reduction algorithm. Some of the families presented in this section are used in Appendix \ref{app:all_cls} to create conservative matrix fields for the corresponding constants.

\subsection{ZigZagZeta - formulas mixing several values of the Riemann zeta function} \label{app:ssec:zzz}
The following continued fractions (Table \ref{table:zzz}) with $a_n$ of degree 5 and $b_n$ of degree 10 were found to have the factorial reduction property. These results led us to discover the ZigZagZeta family, converging to a mix of integer values of the Riemann zeta function, up to half the degree of $b_n$.
\begin{table}[H]
\begin{center}
\def\arraystretch{1.5}
\begin{tabular}{ |c|c|c|} 
\hline
    $a_n$   &   $b_n$   & Formula \\
\hline
    $n^5 + (n+1)^4(n+2)   $     &   $-n^{9}(n+1)$      &   $\frac{1}{\zeta(5)-\zeta(4)+\zeta(3)-\zeta(2)+1}$ \\ 
\hline
     $n^5 + (n+1)^4(n+3)   $     &   $-n^{9}(n+2)$      &   $\frac{32}{32\zeta(5)-48\zeta(4)+56\zeta(3)-60\zeta(2)+61}$ \\ 
\hline
    $n^5 + (n+1)^4(n+4)   $     &   $-n^{9}(n+3)$        &   $\frac{7776}{7776\zeta(5)-14256\zeta(4)+18360\zeta(3)-20700\zeta(2)+21317}$ \\ 

\noalign{\hrule height 2pt}
    \multicolumn{1}{!{\vrule width 2pt}c!{\vrule width 0.5pt}}
        {$n^5 + (n+1)^4(n+1+\alpha)   $} &
        $-n^{9}(n+\alpha)$  &
    \multicolumn{1}{!{\vrule width 0.1pt}c!{\vrule width 2pt}}{$f(\zeta(5), \zeta(4), \zeta(3), \zeta(2))$} \\
\noalign{\hrule height 2pt}

\hline
    $n^4(n+1) + (n+1)^5   $     &   $-n^{9}(n+1)$      &   $\frac{1}{\zeta(4)}$ \\ 
\hline
     $n^4(n+2) + (n+1)^5   $     &   $-n^{9}(n+2)$      &   $\frac{2}{\zeta(4) + \zeta(3)}$  \\ 
\hline
    $n^4(n+3) + (n+1)^5   $     &   $-n^{9}(n+3)$        &   $\frac{6}{2\zeta(4)+3\zeta(3)+\zeta(2)}$  \\ 

\noalign{\hrule height 2pt}
    \multicolumn{1}{!{\vrule width 2pt}c!{\vrule width 0.5pt}}
        {$n^4(n+\alpha) + (n+1)^5   $} &
        $-n^{9}(n+\alpha)$  &
    \multicolumn{1}{!{\vrule width 0.1pt}c!{\vrule width 2pt}}{$g(\zeta(4), \zeta(3), \zeta(2), \alpha)$} \\
\noalign{\hrule height 2pt}

\hline
    $n^4(n+1) + (n+1)^4(n+2)   $     &  
    $-n^{4}(n+1) \times n^{4}(n+1)$      &  
    $\frac{1}{\zeta(4)-\zeta(3)+\zeta(2)-1}$ \\ 
\hline
    $n^4(n+1) + (n+1)^4(n+3)   $     &  
    $-n^{4}(n+1)\times n^4(n+2)  $  &   
    $\frac{16}{16\zeta(4)-24\zeta(3)+28\zeta(2)-29} $ \\ 
\hline
    $n^4(n+2) + (n+1)^4(n+4)   $     &  
    $-n^{4}(n+2)\times n^4(n+3)$     & 
    $\frac{2592}{1296\zeta(4)-1080\zeta(3)+684\zeta(2)-553}$  \\ 

\noalign{\hrule height 2pt}
    \multicolumn{1}{!{\vrule width 2pt}c!{\vrule width 0.5pt}}
        {$\prod\limits_{i=1}^{5}(n+r_i) +  \prod\limits_{i=1}^{5}(n+1+s_i)   $} &
        $-\prod\limits_{i=1}^{5}(n+r_i) \prod\limits_{i=1}^{5}(n+s_i)  $ &
    \multicolumn{1}{!{\vrule width 0.1pt}c!{\vrule width 2pt}}{$t(\zeta(5), \zeta(4), \zeta(3), \zeta(2), \vec{r}, \vec{s})$} \\
\noalign{\hrule height 2pt}

    $2(n^5 + (n+1)^4(n+1+1/2))   $     &  
    $-4n^{9}(n+1/2)$      &  
    $\frac{3}{\pFq{6}{5}{1, 1, 1, 1, 1, 1}{2, 2, 2, 2, 5/2; 1}}$ \\ 
\hline
    $3(n^5 + (n+1)^4(n+1+1/3))   $     &  
    $-9n^{9}(n+1/3)$      &  
    $\frac{4}{\pFq{6}{5}{1, 1, 1, 1, 1, 1}{2, 2, 2, 2, 7/3; 1}}$ \\ 
\hline
    $4(n^5 + (n+1)^4(n+1+1/4))   $     &  
    $-16n^{9}(n+1/4)$      &  
    $\frac{5}{\pFq{6}{5}{1, 1, 1, 1, 1, 1}{2, 2, 2, 2, 9/4; 1}}$  \\ 
\hline

\noalign{\hrule height 2pt}
    \multicolumn{1}{!{\vrule width 2pt}c!{\vrule width 0.5pt}}
        {$C(n^5+(n+1)^4(n+1+1/C)$} &
        $-C^2n^9(n+1/C)$ &
    \multicolumn{1}{!{\vrule width 0.1pt}c!{\vrule width 2pt}}{$\frac{C+1}{\pFq{6}{5}{1, 1, 1, 1, 1, 1}{2, 2, 2, 2, (2C+1)/C; 1}}$} \\
\noalign{\hrule height 2pt}

\end{tabular}
\end{center}
\vspace{-0.5cm}
\caption{\textbf{Results for degree 5 polynomial continued fractions and their generalizations to ZigZagZeta families.} 
These parametric families were manually generalized based on examples discovered by the BOINC community, executing the Distributed Factorial Reduction algorithm. 
The function $f$ that appears in the general form is defined recursively (see below). 
These results can be converted to infinite sums \cite{CMF} using Euler's continued fraction formula and the decomposition $b_n = h_1(n) \cdot h_2(n)$, $a_n = h_1(n) + h_2(n+1)$. This method also provides a closed form for the functions $g$ and $t$.
Similar families exist for other degrees of polynomial continued fractions.}
\vspace{-0.2cm}
\label{table:zzz}
\end{table}

For the case of $a_n = n^s + (n+1)^{s-1}(n+1+R)$ and $b_n=-n^{2s-1}(n+R)$ with values $s\ge2$ and $R \in \mathbb{Q}$ (Eq. \ref{eq:zzz_general}), we denote the limit of the continued fraction by $1/\hat\zeta(s,R)$.
It was found by Wolfgang Berndt that for $R \in \mathbb{N}$, $\hat\zeta(s,R)$ can be calculated by the recursion
$$\hat\zeta(s,R) = \hat\zeta(s,R-1)-\frac{1}{R} \hat\zeta(s-1,R),$$
using initial conditions $\hat\zeta(2,R) =\sum_{j=R+1}^{\infty} {1/j^2}$ and $\hat\zeta(s,0) = \zeta(s)$.

\newpage
\subsection{Infinite family of continued fractions for $\zeta(2)$} \label{app:ssec:zeta2_family}
The following is an infinite family of continued fractions that converge to a linear fractional transformation of $\zeta(2)$, and serves as a basis for the conservative matrix field of $\zeta(2)$ shown in Appendix \ref{app:all_cls}.
\begin{table}[H]
\begin{center}
\def\arraystretch{1.5}
\begin{tabular}{ |l|l|l|c| } 
\hline
    \textbf{$\alpha$} & \textbf{$a_n$} & \textbf{$b_n$} & \textbf{Formula} \\
\hline
    1   &    $n^2 + (n+1)^2$  &  $-n^4$      &   $\frac{1}{\zeta(2)}$        \\
\hline
    2   &    $n^2 + (n+1)^2 + 2$       &  $-n^4$     &   $\frac{1}{2-\zeta(2)}$        \\
\hline
    3   &    $n^2 + (n+1)^2 + 6$      &  $-n^4$     &    $\frac{2}{2\zeta(2)-3}$       \\
\hline
    4   &    $n^2 + (n+1)^2 + 12$       &  $-n^4$     &   $\frac{18}{31-18\zeta(2)}$        \\
\hline 
    \multicolumn{4}{|c|}{...}\\
\noalign{\hrule height 2pt}
    \multicolumn{1}{!{\vrule width 2pt}c!{\vrule width 0.5pt}}{$\alpha$} &
    $n^2 + (n+1)^2 + \alpha(\alpha-1)$  &
    $-n^4$      &
    \multicolumn{1}{!{\vrule width 0.1pt}c!{\vrule width 2pt}}{$\frac{1}{2\Phi(-1, 2, \alpha)}$} 
    \\
\noalign{\hrule height 2pt}
\end{tabular}
\end{center}
\vspace{-0.3cm}
\caption{\textbf{An infinite family of $\zeta(2)$ formulas}. These formulas for $\zeta(2)$ can be expressed through the Lerch zeta function $\Phi(z,s,\alpha)=\sum_{n=0}^{\infty} {\frac{z^n}{(n+\alpha)^s}}$ for $z=-1$, $s=2$, and integer values of $\alpha$. The same formula also holds for non-integer values of $\alpha$.
Using non-integer rational values of $\alpha$ yields formulas that involve more constants other than $\zeta(2)$.
}
\label{table:zeta2_ints}
\end{table}

We observed that the limit of the continued fractions can be  expressed in terms of the Lerch zeta function $\Phi(-1,2,\alpha)$. Substituting non-integer rational values in $\alpha$ gives rise to constants different from $\zeta(2)$. Specifically, this parametric family generalizes several formulas for Catalan's constant found by the first algorithm of the Ramanujan Machine project \cite{TRM,TRM_RESULTS} and by more recent algorithms \cite{naccache2022catalan} that followed on the Ramanujan Machine work.

The family presented here is remarkably similar to the one of $\zeta(3)$ (see Fig. \ref{fig:lattice_motivation}), as both can be expressed through $a_n = n^d + (n+1)^d  + c(n^{d-2} + (n+1)^{d-2})$ and $b_n = -n^{2d}$ for some rational value of $c$. This pattern did not give rise to other infinite families of continued fractions for degrees $d$ higher than 3, but it did produce some sporadic results that we describe in the next subsection.

\subsection{Results for different values of the Riemann zeta function} \label{app:ssec:mixed_zetas}
In an attempt to generalize %to follow
the formulas found for $\zeta(2)$ (Table \ref{table:zeta2_ints}) and $\zeta(3)$ (Fig. \ref{fig:lattice_motivation}) to higher $\zeta$ values, we 
executed the Distributed Factorial Reduction algorithm over a specially constructed sub-space of polynomial continued fractions.
Specifically, we searched over the following parametric family with integer parameters $B,c_d,c_{d-2},c_{d-4}...$ (down to $c_1$ or $c_0$ depending on the parity of $d$):
\begin{equation} \label{eq:sigma_poly_struct}
    a_n = c_d \sigma(n,d) + c_{d-2} \sigma(n,d-2) + 
    c_{d-4} \sigma(n,d-4)+\dots  \qquad b_n = -B n^{2d},
\end{equation}
% $$a_n = c_0 + c_1 \sigma(n,1) + c_2 \sigma(n,2) + ... + c_d \sigma(n,d) \qquad b_n = -B n^{2d},$$
defining $\sigma(n, d) = n^d + (n+1)^{d}$.
%Part of the resulting formulas are members of the infinite families described in Table \ref{table:zzz} and in the other sections. 
The resulting formulas do not belong to the infinite families described in other sections and are instead sporadic formulas (Table \ref{table:sporadic_zetas}). It is currently unknown whether any of the sporadic formulas can be generalized to form a new infinite family. 

\begin{table}
\begin{center}
\def\arraystretch{1.5}
\begin{tabular}{ |c|c|c| } 
\hline
    $a_n$   &   $b_n$   & Formula \\
\hline
    $n^4 + (n+1)^4 + 2(n^2 + (n+1)^2)  $     &   
    $-n^{8}$      &   $\frac{1}{-\zeta(4)-2\zeta(2)+8}$ \\ 
\hline
    $n^5 + (n+1)^5 + 6(n^3 + (n+1)^3)                   $     &   $-n^{10}$      &   $\frac{2}{2\zeta(5)+6\zeta(3)-9}$ \\ 
\hline
    $n^5 + (n+1)^5 + 6(n^3 + (n+1)^3) - 4(2n+1)         $     &   $-n^{10}$      &   $\frac{2}{2\zeta(5)-2\zeta(3)-1}$ \\ 
\hline
    $n^5 + (n+1)^5 + 16(n^3 + (n+1)^3) - 4(2n+1)         $     &   $-n^{10}$      &   $\frac{64}{64\zeta(5)+176\zeta(3)-273}$ \\ 
\hline
    $8 (n^5 + (n+1)^5) + 20(n^3 + (n+1)^3) - 5(2n+1)         $     &   $-64n^{10}$      &   $22.8410...$ \\

\hline
    $8 (n^5 + (n+1)^5) - 15(n^3 + (n+1)^3) + 9(2n+1)$ &
    $-64n^{10}$  &
    $1.20426...$
    \\
\hline  
    $8 (n^5 + (n+1)^5) - 12(n^3 + (n+1)^3) + 7(2n+1)         $ &
    $-64n^{10}$  &
    $2.45174...$
    \\
\hline
    $n^7 + (n+1)^7 + 8(n^5 + (n+1)^5) - 8(n^3 + (n+1)^3) + 4(2n+1)         $     &   $-n^{14}$      &   $\frac{1}{\zeta(7)-4\zeta(3)+4}$ \\ 
\hline

\end{tabular}
\end{center}
\vspace{-0.3cm}
\caption{\textbf{Sporadic results for higher values ($\ge4$) of the Riemann zeta function}. All the continued fractions in this table exhibit factorial reduction. Lines 5 to 7 are only formula \textit{candidates} as we did not find a closed-form expression for them, despite their recursion formulas being strikingly similar to the others. 
We would expect them to yield a formula involving constants from the same level of the hierarchy as $\zeta(5)$.}
\label{table:sporadic_zetas}
\end{table}

\subsection{Infinite family of continued fractions for values of the Polylogarithm function}
Attempting to discover new formulas that converge to expressions involving $\zeta(4)$, we executed the Distributed Factorial Reduction algorithm scanning continued fractions of degree 4.
While the search did not yield many formulas explicitly including $\zeta(4)$, numerous formulas presenting constants from the same level of the hierarchy emerged. 
Among these results, a subset of continued fractions yielded values of the fourth order Polylogarithm function $Li_4(x)$.
Subsequently, these findings were rigorously proven using Euler's continued fraction formula and generalized to any order of the Polylogarithm function.
This generalized continued fraction family is defined through the polynomials:
$$
    a_n = n^d + c (n+1)^d \qquad b_n = -cn^{2d}.
$$
The resulting continued fractions converge to $\mathrm{Li}_d (1/c)$.

Inspired by this parametric family, we found that the ZigZagZeta family can be similarly expanded. Indeed, adding a parameter $c$ to the general formula from Table \ref{table:zzz} creates 
the following continued fractions that have the factorial reduction property:
$$a_n = \prod_{i=1}^{d}(n+r_i) + c\prod_{i=1}^{d}(n+1+s_i) $$
$$b_n = -c\prod_{i=1}^{d}(n+r_i) \times  \prod_{i=1}^{d}(n+s_i). $$

\section{Discovered conservative matrix fields}  \label{app:all_cls}

This section presents examples of conservative matrix fields for several important mathematical constants.

\subsection{The conservative matrix field of $e$} \label{app:ssec:e_cmf}
The Ramanujan Machine \cite{TRM} generated hundreds of continued fraction formulas for $e$. Generalizing these formulas, we constructed parametric families of continued fractions, which led us to find the following conservative matrix field:
$$
M_X = \begin{pmatrix}
    0   &   x+1 \\
    1   &   -(x+y+1)
\end{pmatrix}, 
M_Y = \begin{pmatrix}
    -1  &   x+1 \\
    1   &   -(x+y+2)
\end{pmatrix}
$$
This matrix field can be generated through the degree 1 analytical construction of matrix fields presented in Appendix \ref{app:CMF_families} by setting $\vec{c} = (0,-1,-1,0)$.
By following trajectories in other quadrants, this conservative matrix field % these constants 
can also generate sequences that converge to linear fractional transformations of the Euler–Gompertz constant and to other values of the ${}_1 F_1$ hyper-geometric function.

\subsection{The conservative matrix field of $\pi$}
Through an identical process to the $e$ matrix field, we created the following matrix field for $\pi$:
$$
M_X = \begin{pmatrix}
    0   &   -(2x+1)x \\
    1   &   y+3x+2
\end{pmatrix}, 
M_Y = \begin{pmatrix}
    y-x  &   -(2x+1)x \\
    1   &   2x+2y+1
\end{pmatrix}
$$
This matrix field was later identified as a member of the degree 1 analytical matrix field presented in Appendix \ref{app:CMF_families} by setting $\vec{c} = (1,2,0,-1)$.

\subsection{The conservative matrix field of $\zeta(2)$}
The infinite family of continued fractions shown in Table \ref{table:zeta2_ints} enabled us to construct a conservative matrix field that converges to linear fractional transformations of $\zeta(2)$. 
The matrices that define the matrix field are:
$$
M_X = \begin{pmatrix}
%    x^2 + (x+1)^2 +y(y-1)       & -x^2 \\
%    (x+1)^2                         & 0
    0  & -x^2 \\
    (x+1)^2                         & x^2 + (x+1)^2 +y(y-1)
\end{pmatrix}
$$
$$
M_Y = \begin{pmatrix}
%    x^2 +xy+y^2/2              & -x^2 \\
%    x^2                        & -x^2 + xy -y^2/2s
    -x^2 + xy -y^2/2 & -x^2 \\
    x^2                        & x^2 +xy+y^2/2
\end{pmatrix}
$$

Section \ref{ssec:rational_shifts} discusses the effect of applying a rational shift to the $x$ or $y$ variables. 
Applying such a shift in this case leads to a conservative matrix field that converges to expressions involving Catalan's constant $G$ and other constants. 
Naccache et al. \cite{naccache2022catalan,naccache2023balkans} have recently generated an infinite family of formulas for the Catalan constant
by a new approach following the Ramanujan Machine.
The conservative matrix field of $\zeta(2)$ (with its extra dimensions in the next section) seems to generate these
% the same
infinite families and generalize them to higher-dimensional parametric spaces. 

\section{Multidimensional conservative matrix field of degree 2} \label{app:multidimensional_cmf}

The definition of a conservative matrix field can be modified to allow for grids of dimension $d > 2$, specified by $d$ matrices that satisfy the pairwise conservative property:
\begin{equation} \label{multidimensional_conservative_condition}
\begin{split}
    M_{U_i} (u_0, ...,u_i,.., u_j, ..., u_d) M_{U_j} (u_0, ...,u_i+1,.., u_j, ..., u_d) = \\
    M_{U_j} (u_0, ...,u_i,.., u_j, ..., u_d) M_{U_i} (u_0, ...,u_i,.., u_j+1, ..., u_d) 
\end{split}
\end{equation}
Multidimensional matrix fields support more trajectories, and with them a possibility for faster converging sequences and larger irrationality measures. 

We formulated a set of four matrices that satisfy the generalized conservative property, using second degree polynomials as matrix entries. %with polynomials of degree 2. 
%, that contains the matrix field of $\zeta(2)$ as plain in it. 
Each of these four matrices can be constructed through a single matrix (with an arbitrary rational $C$):
$$
M = \begin{pmatrix}
    C(xy + xz + xw + yz + yw + zw)  &   -1 \\
    C^2xyzw                        &   0
\end{pmatrix}
 + Cx^2 \cdot \mathbb{I},
$$
which enables composing the 4D matrix field:
\begin{equation}
\begin{split}
    M_X(x,y,z,w) = M(x, y, z, w) \\
    M_Y(x,y,z,w) = M(y, z, w, x) \\
    M_Z(x,y,z,w) = M(z, w, x, y) \\
    M_W(x,y,z,w) = M(w, x, y, z). 
\end{split}
\end{equation}
Below, we examine the case of $C=1$.

Surprisingly, this 4D matrix field can generate continued fractions from the ZigZagZeta family (Appendix \ref{app:ssec:zzz}). To achieve this intriguing result, a specific transformation must be applied to the matrices, forcing them into a form that represents a continued fraction (as presented in Eq. \ref{eq:pcf_to_mat_mult}). This transformation involves taking a co-boundary on the matrices, which can be achieved by using a matrix denoted as $U(x,y,z,w)$ and applying it to the matrices in each direction according to the following rule: $M_s \to U^{-1} M_s U(s \to s+1)$, where $s$ represents the direction index, and $U$ is defined as:
$$U = \begin{pmatrix}
    1  &   xy + xz + xw + yz + yw + zw + x^2 \\
    0  &   xyzw
\end{pmatrix}.$$
Along the $x$ axis, such a transformation retrieves the ZigZagZeta family of formulas:
$$M_x = \begin{pmatrix}
    0  &   -(x+1)(x+y)(x+z)(x+w) \\
    1  &   (x+1)^2 + (x+1)(x + y + z + w) + (yz + zw + wy)
\end{pmatrix}.$$
This $M_x$ represents the polynomial continued fraction with $a_n=(x+1)^2 + (x+1)(x + y + z + w) + (yz + zw + wy)$ and $b_n = -(x+1)(x+y)(x+z)(x+w)$, exactly matching the ZigZagZeta family shown in Appendix \ref{app:ssec:zzz}.

\section{Methods and algorithms for the conservative matrix fields} \label{app:cmf_algos}

The conservative matrix fields are novel mathematical structures with various applications, such as the generation of infinitely many %second order linear 
recurrence formulas that correspond to fundamental constants. This new mathematical structure calls for developing new algorithmic tools tailored to its unique properties.
This section presents ways to extract optimal trajectories in matrix fields and explains how to transform each trajectory to a continued fraction.

\subsection{Algorithms to identify optimal trajectories for maximal irrationality measures} \label{app:ssec:cmf_traj_algos}
Different trajectories on the conservative matrix field offer different values of $\delta$. 
In some cases, as depicted in Fig. \ref{fig:zeta3_delta}, the trajectory $x=y$ exhibits the largest $\delta$. 
But this phenomenon is not consistent across other matrix fields, leading to a new challenge in identifying the optimal trajectory on a matrix field. 
The challenge becomes more significant when dealing with higher-dimension matrix fields (Appendix \ref{app:multidimensional_cmf}), or when attempting to search a large number of matrix fields for trajectories that yield positive $\delta$. 

To search for the optimal trajectory, the first algorithm that we apply is a version of gradient decent. We begin at the origin and calculate $\delta(x,y)$ in neighboring locations 
using Eq. \ref{eq:dioph_CMF}. We then choose the step that maximizes $\delta(x,y)$, and repeat the process.
As an example, Fig. \ref{fig:zeta3_delta}(b) presents the trajectory found by this algorithm in black. This algorithm can often miss the optimal trajectory when the function $\delta(x,y)$ is not smooth enough. 
Various generalizations of gradient-descent algorithms can be directly applied to address this issue and yield better results.

The second algorithm that we apply uses linear least squares (LLS) to find an optimal trajectory.
For any trajectory initiated at coordinate $(1,1)$, the endpoint after $n$ steps would be a coordinate $(x,y)$ where $x+y-2=n$. We analyze the $\delta$ at each coordinate satisfying this condition to determine the best achievable $\delta$ at the $n$th step. We then repeat the process for increasing $n$ values and use LLS to fit a linear path of optimal $\delta$. 
For the case depicted in Fig. \ref{fig:zeta3_delta}(b), the trajectory obtained via this method provides  
%closest irrationality measure to the optimal $\delta=-0.4741$).
a larger irrationality measure relative to gradient descent methods, reaching $\delta=-0.4741$ for this constant that combines $\zeta(3)$ and $\pi^3$ and is not known to be rational or irrational.

\subsection{Equivalence between continued fractions and conservative matrix fields} \label{app:ssec:pcf_to_cmf_equivalence}
%For the $\pi$ conservative matrix field, 
The matrix attained by choosing the $x=y$ trajectory is 
$$M_n(n) = M_x(n,n) M_y(n+1,n).$$
This matrix does not generally have a continued fraction form (like $M_x$ in Eq. \ref{eq:pcf_to_mat_mult}). However, it is always possible to transform this matrix to take the form of a continued fraction.
In fact, there are certain transformations that convert an arbitrary sequence of matrix multiplications to a continued fraction form \cite{CMF}\cite{ESMA}. i.e., it is always possible to extract effective polynomials $a_n,b_n$ for any sequence of $2\times2$ matrix multiplications.
For example, calculating the conservative matrix field of $\pi$ along the trajectory $x=y$ creates the following formula for $\pi$: 
\begin{equation}
\frac{10}{\pi-4} = 
-12 + \cfrac{-1}
   {-238 + \cfrac{-16}
       {-968 + \cfrac{-81}
           {\ddots + \cfrac{-n^2(2n+1)^2(4n-3)(4n+5)}
               {-2(4n+3)(6n^2+9n+2) + \ddots}}}}
\end{equation}

The conversion from the polynomial continued fraction to a matrix multiplication changes the initial conditions by a linear fractional transformation with integer coefficients, which does not affect the properties of the sequence.
This behavior is part of a more general property of conservative matrix fields.
We can always change the initial condition of a trajectory (the position or initial matrix) and the resulting constant only changes by a linear fractional transformation with integer coefficients. We refer to such cases as converging to the ``same'' constant.

An implementation of this algorithm can be found in our git repository 
\url{https://github.com/RamanujanMachine/ResearchTools}

\section{Using families of polynomial continued fractions to derive non-trivial irrationality measures for $\zeta(5)$}  \label{app:ZZZ_lattice}
\begin{wrapfigure}{r}{6cm}
    \vspace{-0.7cm}
    \centering
    \includegraphics[width=6cm]{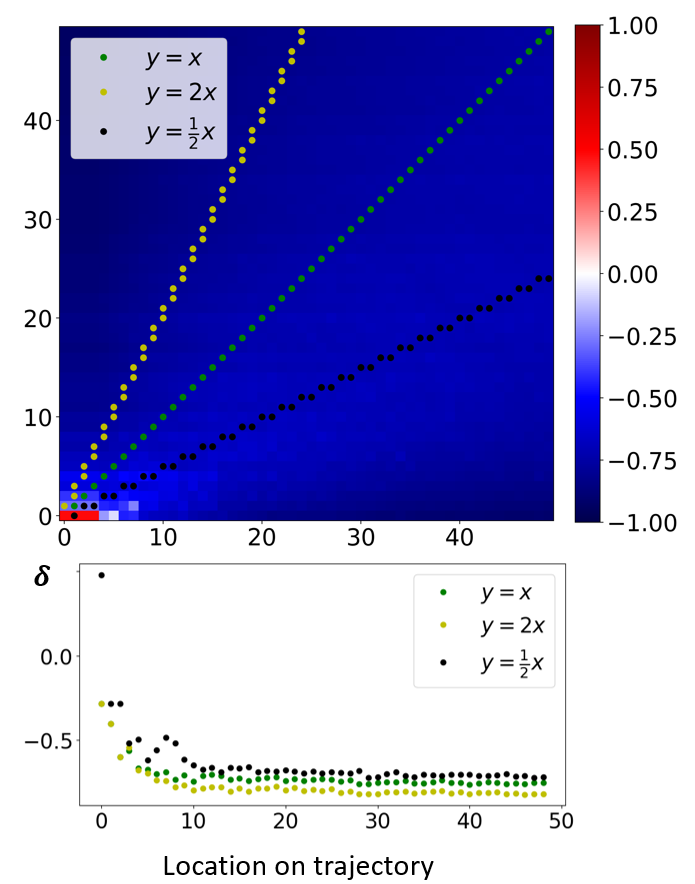}
    \caption{\textbf{Example delta approximations for $\zeta(5)$}, acquired by taking different trajectories on a 2D space of ZigZagZeta continued fraction formulas.
    }
    \label{fig:ZZZ_lattice}
    \vspace{-2cm}
\end{wrapfigure}
Table \ref{table:zzz} presented several ZigZagZeta families that converge to expressions involving multiple constants. In this section, we combine these formulas to generate sequences converging to a single constant, by linear combinations that eliminate unwanted constants.

To exemplify this concept, consider a special case of Eq. \ref{eq:zzz_general}, where we substitute $R=1$.
Two members of this family are:
$$\mathrm{PCF} [s=4,R=1] = \frac{1}{\hat\zeta(4,1)} = \frac{1}{\zeta(4) - \zeta(3) + \zeta(2) - 1}, $$
$$\mathrm{PCF} [s=5,R=1] = \frac{1}{\hat\zeta(5,1)} = \frac{1}{\zeta(5) - \zeta(4) + \zeta(3) - \zeta(2) + 1}. $$
Summing the inverse of each of these polynomial continued fractions yields a sequence that converges to $\zeta(5)$:
$$
\zeta(5) = \frac{1}{\mathrm{PCF} [s=5,R=1]} + \frac{1}{\mathrm{PCF} [s=4,R=1]}.
$$
By taking the convergents of the continued fractions, this formula produces a rational sequence converging to $\zeta(5)$.

\vspace{0.8cm}
We can directly generalize this approach by taking linear combinations of formulas with additional values of $R$, producing a multi-dimensional infinite family of formulas of any value of the Riemann zeta function.
%As the inverse of all elements of the ZigZagZeta family converges to linear combinations of various values of $\zeta$, we can repeat this procedure indefinitely by selecting multiple members of the family, generating a set of linear equations from them, and solving for the desired $\zeta$ value. The outcome of this process yields a formula for $\zeta$, consisting of different continued fractions from that family.
For example, we take the following formulas with the same $R$ for all powers $s$ up to 5:
\begin{equation}
\left\{
\begin{matrix}
\begin{array}{l}
\cfrac{1}{\mathrm{PCF} [s=2,R]} = \hat{\zeta}(2,R) = \zeta(2) + \alpha_2 \\
\cfrac{1}{\mathrm{PCF} [s=3,R]} = \hat{\zeta}(3,R) = \zeta(3) + \beta_3 \zeta(2) + \alpha_3 \\
\cfrac{1}{\mathrm{PCF} [s=4,R]} = \hat{\zeta}(4,R) = \zeta(4) + \gamma_4 \zeta(3) + \beta_4 \zeta(2) + \alpha_4 \\ 
\cfrac{1}{\mathrm{PCF} [s=5,R]} = \hat{\zeta}(5,R) = \zeta(5) + \delta_5 \zeta(4) + \gamma_5 \zeta(3) + \beta_5 \zeta(2) + \alpha_5.
\end{array}
\end{matrix}
\right.
\end{equation}
These linear equations form a matrix that can be inverted to find the coefficients $c_i$, providing an explicit expression for
$
\zeta(5) = c_2 \cdot \hat{\zeta}(2,R) + c_3 \cdot \hat{\zeta}(3,R) + c_4 \cdot \hat{\zeta}(4,R) + c_5 \cdot \hat{\zeta}(5,R).
%\zeta(5) = \frac{c_2}{\mathrm{PCF} [d=2,k]} + \frac{c_3}{\mathrm{PCF} [d=3,k]} + \frac{c_4}{\mathrm{PCF} [d=4,k]} +\frac{c_5}{\mathrm{PCF} [d=5,k]}
$

Substituting the corresponding continued fraction into each $\hat{\zeta}$ yields an infinite family of rational sequences that all converge to $\zeta(5)$. We place these sequences as lines of a 2D grid
and 
%use the approach suggested in Section \ref{ssec:irrationality_from_cmfs}, we can 
search for optimal sequences along different trajectories on it, assessing the irrationality measure provided by each trajectory. Following this procedure for the specific 2D grid presented here already provides a non-trivial irrationality measure for $\zeta(5)$ of $\delta \approx -0.717$. Applying this procedure on the much richer multi-dimensional infinite family of formulas that we found could help optimize $\delta$, potentially increasing it to above 0, and thus providing a path for proving the irrationality of $\zeta(5)$.

\newpage
\section{Analytical construction of conservative matrix fields}
\label{app:CMF_families}

% Short introduction - what is CMF why do we need it, ref to other paper
% One way to construct them : linear + quadratic condition.
% Examples: 
%   1. e, pi, zeta(3), zeta(2) - every direction converges to same place (?)
%   2. there are many other construction - deg 1, deg 2. Add ref to the right section about convergence in different directions

%The conservative matrix field is, at its core, a collection of matrices that satisfy the \textit{the conservative property} (Eq. \ref{eq:cl_conservative_condition}). For a two-dimensional matrix field, this property can be expressed as:

%$$M_X(x,y) \cdot M_Y(x+1,y) = M_Y(x,y) \cdot M_X(x,y+1)$$

In other sections of this paper (Section \ref{ssec:CMF}, Appendix \ref{app:all_cls}) we have derived matrices $M_X$ and $M_Y$ that define a conservative matrix field by generalizing an infinite family of continued fractions. In this section, we present another approach, which involves finding a general expression for the elements of $M_X$ and $M_Y$ that satisfy the conservative property. A detailed review of this process is presented in \cite{CMF}. The outcome of this analysis is the realization that an infinite family of matrix fields can be constructed using two functions, $f(x,y)$ and $\bar{f}(x,y)$, that satisfy the following two conditions:
\begin{itemize}
    \item \textbf{Linear condition: } 
    $f(x,y)-f(x+1,y-1) = \bar{f}(x+1,y)- \bar{f}(x,y-1)$
    \item \textbf{Quadratic condition: } 
    $ (f\bar{f})(x,y) + (f\bar{f})(0,0) = (f\bar{f})(x,0) + (f\bar{f})(0,y)$
\end{itemize} 
Note that the linear condition is a modified version of the defining property of Wilf--Zeilberger pairs \cite{A_eq_B_Zeilberger}. 

Given such $f(x,y)$ and $\bar{f}(x,y)$, we define:
\begin{align*}
&a(x,y) =f(x,y)-\bar{f}(x+1,y), \\
&b(x) = (f\bar{f})(x,0) - (f\bar{f})(0,0).
\end{align*}
We can then obtain the matrices $M_X$ and $M_Y$ as follows:
\begin{equation}
    \begin{matrix}
        M_X(x,y) = \begin{pmatrix}
            0 & b(x)  \\
            1 & a(x, y)
        \end{pmatrix}, &
        M_Y(x,y) = \begin{pmatrix}
            \bar{f}(x,y) & b(x)  \\
            1 & f(x, y)
        \end{pmatrix}
    \end{matrix}
\end{equation}

By setting $f(x,y)$ and $\bar{f}(x,y)$ to be polynomials of the same degree, we were able to find closed-form solutions for the linear and quadratic conditions for $f, \bar{f}$ of degrees 1 and 2. The solution takes the form of a parametric family of conservative matrix fields.
For $f, \bar{f}$ of degree 1, each conservative matrix field is defined by four parameters $\vec{c} = (c_0,c_1,c_2,c_3)$, and the solutions are of the form:
\begin{equation} \label{eq:deg1_cmf}
\begin{split}
f(x,y) = c_0 + c_1(x + y) \\
\bar{f}(x,y) = c_2 + c_3(x - y)
\end{split}
\end{equation}
The horizontal trajectories in this matrix field correspond to polynomial continued fractions with partial numerator $b_n$ and denominator $a_n$ of degrees at most 2 and 1, respectively. For such continued fractions, the limit can be computed in terms of the hypergeometric functions ${}_1 \mathrm{F}_1$ and ${}_2 \mathrm{F}_1$ in \cite[\S 49]{Perron2}. 

In particular, the degree 1 matrix field can produce exponential, logarithms, and algebraic numbers as their limit, among other values. We have identified families of $\vec{c}$ parameters that create matrix fields converging to predictable limits:
$$ \vec{c}=(0,k,l,0) \to e^{\frac{l}{k}} \qquad
 \vec{c}=(0,k,0,m) \to \ln\left(\frac{k+m}{k}\right) \qquad
 \vec{c}=(0,k,l,m) \to \left(\frac{k+m}{k}\right)^{\frac{l}{m}} $$
These formulas have been empirically verified (each tested numerically along the $x = y$ diagonal). The obtained constants coincide with the limits along horizontal directions of each of the corresponding matrix fields, which can be computed in terms of the hypergeometric functions ${}_1\mathrm{F}_1$ and ${}_2\mathrm{F}_1$ (see \cite[\S 48, 49]{Perron2}).

Interestingly, the parametrization of Eq. \ref{eq:deg1_cmf} encapsulates various properties of conservative matrix fields, such as the shifts to the initial location of different trajectories as described in Sec. \ref{ssec:rational_shifts}. 
The parameters $c_0$ and $c_2$ can be thought of as shifts along the $x+y$ and $x-y$ respectively, by $\frac{c_0}{c_1}$ and $\frac{c_2}{c_3}$ (assuming $c_1\neq0$ or $c_3\neq0$). Since we take the limit along the $x=y$ axis, a shift by an integer along it is equivalent to changing the starting point of the trajectory, which only introduces a linear fractional transform on the limit, but does not change the fundamental constant underlying the limit.

\newpage
One consequence of the above is that the following matrix fields converge to the same constant (up to a linear fractional transform), 
for any integer $N$:
$$ (c_0,c_1,c_2,c_3) \Longleftrightarrow (c_0+Nc_1,c_1,c_2,c_3) $$
An example of this effect can be seen in the entry $\vec{c}=(2,2,0,1)$ in Table \ref{table:ir_proofs_with_cmfs}, which converges to a linear fractional transform of $\ln\left(\frac{3}{2}  \right)$, the same constant as in the equation above for $k=2, m=1$, i.e., $\vec{c}=(0,2,0,1)$.
%. According to the claim above, $\vec{c}=(2,2,0,1)$ should converge to the same fundamental irrational number as $\vec{c}=(0,2,0,1)$. Substituting $k=2, m=1$ in our second formula above, we indeed get $\ln\left(\frac{3}{2}  \right)$ as predicted. 
Another example is the table entry $\vec{c}=(3,3,2,4)$, which converges to a linear fractional transform of $\sqrt{21}$, the same constant as in the equation above for $k=3, l=2, m=4$, i.e., $\vec{c}=(0,3,2,4)$.

%, so if we take the third formula above, substitute $k=3, l=2 m=4$, we expect to get $(7/3)^{1/2}$ - which again fits the equality with see in Table \ref{table:ir_proofs_with_cmfs}. 
% We see the fundamental irrationality in the numerical result is $\sqrt{21}$, which makes - as we're only interested in results up to a linear fractional transform, so we can multiply by any integer (3 in this case), and get $\left(3^2\cdot\frac{7}{3}\right)^{\frac{1}{2}} = \sqrt{21}$.

\begin{comment}
    Something doesn't work here for me.
    Lets take c=(0,1,0,2). This CMF converges to ln(3) (found numerically 0.82047845325367478722848033147221400122527211). 
    (Using Carlos' document from the drive)
    When I try to apply it here, I find:
    f  = (x+y)
    f' = 2(x-y)

    a(x,0) = x - 2(x+1) = -2 -x
    b(x) = 2x^2
    so:
    a = 0 ; b = 0 ; c = 2
    d = -2 ; e = -1

    alpha and beta are solutions of:
        cz^2 - bz + a = 0
        ---> alpha = beta = 0
    then 
        delta = e^2 + 4c = 9
        sqrt(delta) = -3 (to match the sign of e)
        
    z = 1/2 * (1-e/sqrt(delta)) = 1/2 (1 - 1/3) = 1/3 
    
    gamma = (b+c)z/c + d/sqrt(delta) = 1/3 -2/(-3) = 1

    the limit is ln(3/2)
    https://www.wolframalpha.com/input?i=gamma%282%29%2Fgamma%281%29+*++2F1%280%2C+0%2C+1%2C+1%2F3%29%2F2F1%281%2C+1%2C+2%2C+1%2F3%29
\end{comment}

The degree 1 matrix field captures additional cases that we discovered using algorithmic approaches (Appendix \ref{app:all_cls}). For example, setting $c_1 = (1,2,0,1)$ and transforming the matrix to its balanced form (\cite{endnote_balanced_matrix}) results in the conservative matrix field of $\pi$.

For $f, \bar{f}$ of degree 2, each conservative matrix field is defined by four parameters $(c_0,c_1,c_2,c_3)$, and the solutions are of the form:
$$f(x,y) = (2c_1 + c_2)(c_1 + c_2) - c_3 c_0 - c_3 ((2c_1 + c_2) (x+y) + (c_1 + c_2)(2x + y)) + c_3^2 (2x^2 + 2xy + y^2),$$
$$\bar{f}(x,y) = c_3 (c_0 + c_2 x + c_1 y - c_3  (2x^2 - 2xy + y^2)).$$
One special case is found by setting $\vec{c} = (0,0,0,1)$, yielding the conservative matrix field for $\zeta(2)$ shown in Appendix \ref{app:all_cls}. 

As the degree of $f$ and $\bar{f}$ increases, the linear and quadratic conditions evolve into a larger number of more complicated conditions. We leave this challenge to future works, and present below an instructive special case. The degenerate case in which $f(0,0)=\bar{f}(0,0)=0$ includes exactly three families of $f$ and $\bar{f}$ polynomials of degree 3. Each of these families is defined by two parameters $(c_0,c_1)$:

\begin{itemize}
    \item 
    $f_1(x,y) = -((c_0+c_1 (x+y))(c_0 (x+2 y)+c_1 (x^2+x y+y^2 )))$
    $\bar{f_1}(x,y) = (c_0+c_1 (-x+y))(c_0 (x-2 y)-c_1 (x^2-x y+y^2 )) $
    \item
    $f_2(x,y) = (c_0+c_1 (-x+y))(c_0 (x-2 y)-c_1 (x^2-x y+y^2 ))$
    $\bar{f_2}(x,y) = (c_0+c_1 (-x+y))(c_0 (x-2 y)-c_1 (x^2-x y+y^2 ))$
    \item
    $f_3(x,y) = (x+y)(c_0^2-c_0 c_1 (x-y)-2 c_1^2 (x^2+x y+y^2 ))$
    $\bar{f_3}(x,y) = (c_0+c_1 (x-y))(3 c_0 (x-y)+2 c_1 (x^2-x y+y^2 ))$
\end{itemize}
Choosing $f_1, \bar{f}_1$ and $c=(0,1)$ results in the conservative matrix field for $\zeta(3)$ (Fig. \ref{fig:lattice_motivation}, Eqs. \ref{eq:zeta3_M_X},\ref{eq:zeta3_M_Y}).

In Appendix \ref{app:ir_meas_cmf_family}, we evaluate irrationality measures of the approximations produced by diagonal trajectories inside these families of conservative matrix fields, 
identifying ones that can be used to prove the irrationality of certain constants.

\newpage
\section{Using conservative matrix fields to derive irrationality measures of different constants} \label{app:ir_meas_cmf_family}
The family of conservative matrix fields presented in Appendix \ref{app:CMF_families} generates infinitely many options to extract rational sequences with accelerated convergence that can each help proving the irrationality of a certain constant. Different members of the family converge to different constants, providing a systematic methodology of creating potentially irrationality-proving sequences to different constants. 

By calculating sequences generated from the $x=y$ trajectory of each conservative matrix field, we constructed dozens of irrationality-proving sequences. Table \ref{table:ir_proofs_with_cmfs} shows a set of such sequences and their respective irrationality measures. We note that some of the constants presented here are quite exotic and hard to fit with PSLQ. Thus, we also used inverse-calculators as Wolfram's closed-forms detection to connect these sequences to their matching limits.

\begin{table}[H] 
\begin{center}
\def\arraystretch{1.5}
\begin{tabular}{ |l|l|l|c| } 
\hline
    \textbf{Constant} & 
    \textbf{CMF degree} & 
    \makecell{\textbf{CMF parameters} \\ \textbf{$(c_0,c_1,c_2,c_3)$}} &
    \makecell{\textbf{Numerically} \\ \textbf{optimized $\delta$}} \\
\hline
    $4^{1/3}$ &     1 &    (0, 1, 1, 3) & $0.0938$ \\
\hline
    $14^{1/3}$ &    1 &    (0, 4, 4, 3) & $0.376$ \\
\hline
    $2^{1/3}$ &     1 &    (3, 3, 2, 3) & $0.321$ \\
\hline
    $5^{1/3}$ &     1 &    (0, 5, 1, 3) & $0.405$\\
\hline
    $e$       &     1 &    (0, 2, 2, 0) & $1.0$\\
\hline
    $\sqrt{e}$ &    1 &    (0, 4, 2, 0) & $1.0$\\
\hline
    $e^2$      &    1 &    (0, 1, 2, 0) & $1.0$\\
\hline
    $\ln(2)$   &    1 &    (0, 1, 0, 1) & $0.288$\\
\hline
    $\sqrt{2}$ &    1 &    (0, 4, 2, 4) & $1.0$ \\
\hline
    $\sqrt{3}$ &    1 &    (0, 1, 1, 2) & $1.0$\\
\hline
    $\sqrt{5}$ &    1 &    (0, 5, 2, 4) & $1.0$    \\
\hline
    $\sqrt{6}$ &    1 &    (0, 4, 1, 2) & $1.0$    \\
% Now some funky examples
\hline
    $\cfrac{1}{4}\left(-5\sqrt{21} + \sqrt{2(21+5\sqrt{21}} \right)$ &
     1 &    (0, 3, 1, 4) & $0.0137$ \\ 
\hline
    $\cfrac{9}{\sqrt[4]{3\cdot7^3} - 3}$ &
     1 &    (0, 3, 3, 4) & $0.0098$ \\ 
\hline
    $\cfrac{1}{3} \left( -5 -2 \sqrt[3]{14} + \sqrt[3]{14^2} \right)$ &
     1 &    (0, 4, 1, 3) & $0.3621$ \\ 
\hline
    $\cfrac{5 - 12 \log{3} + \log{4096}}{\log{27/8} - 1}$ &
     1 &    (2, 2, 0, 1) & $0.4041$\\ 
\hline
    $\cfrac{1}{25} \left(21 \sqrt{21} - 19\right)$ &
     1 &    (3, 3, 2, 4) & $0.9830$\\ 
\hline
    $\zeta(2)$  &  2 &    (0, 0, 0, 1) &  $0.0988$
     \\
\hline
    $\zeta(3)$  &  3 &    (0, 1) using $f_1, \bar{f}_1$ &     $0.0833$\\ 
\hline
\end{tabular}
\end{center}
\vspace{-0.5cm}
\caption{\textbf{Irrationality measures for various constants}, using the $x=y$ trajectory in different conservative matrix fields.}
\label{table:ir_proofs_with_cmfs}
\end{table}

%\subfile{appendix}

\end{document}